\documentclass[conference]{IEEEtran}
\IEEEoverridecommandlockouts
\usepackage{cite}
\usepackage{algorithmicx, algpseudocode}
\usepackage{amsmath}
\usepackage{amsfonts}
\usepackage{amsbsy}
\usepackage{amsthm}
\usepackage{multirow}
\usepackage{booktabs}
\usepackage[mathscr]{eucal} 
\usepackage{mathtools}
\usepackage{verbatim}
\usepackage{tabularx}
\usepackage{enumitem}
\usepackage{array}
\usepackage{soul}
\usepackage{dsfont}
\usepackage{url}
\usepackage{bbm}
\usepackage{pifont}
\usepackage{xcolor,colortbl}
\usepackage[inkscapeformat=png]{svg}
\usepackage{makecell}
\usepackage{contour}
\usepackage{ulem}
\usepackage{tcolorbox}
\contourlength{0.8pt}

\newcommand{\myuline}[1]{%
  \uline{\phantom{#1}}%
  \llap{\contour{white}{#1}}%
}

\usepackage{xspace}

\definecolor{newgreen}{rgb}{0.0, 0.5, 0.0}
\definecolor{newred}{rgb}{0.81,0.1,0.26}
\definecolor{babyblue}{rgb}{0.54, 0.81, 0.94}
\definecolor{negcol}{rgb}{0.96, 0.76, 0.76}
\definecolor{poscol}{rgb}{0.63, 0.79, 0.95}
\definecolor{jongha}{rgb}{0.66, 0.89, 0.63}
\definecolor{sunwoogreentwo}{rgb}{0.53, 0.81, 0.98}
\definecolor{dodgerblue}{rgb}{0.12, 0.56, 1.0}
\definecolor{crimson}{rgb}{0.86, 0.08, 0.24}
\definecolor{limegreen}{rgb}{0.2, 0.8, 0.2}
\definecolor{backredcolor}{rgb}{0.91, 0.45, 0.32}

\newcommand{\goat}

\let\oldnl\nl
\newcommand{\nlnonumber}{\renewcommand{\nl}{\let\nl\oldnl}}

\def\BibTeX{{\rm B\kern-.05em{\sc i\kern-.025em b}\kern-.08em
    T\kern-.1667em\lower.7ex\hbox{E}\kern-.125emX}}

\begin{document}

    \title{Simple yet Effective Node Property Prediction on Edge Streams under Distribution Shifts}
    
	
    \author{\IEEEauthorblockN{Jongha Lee,
    Taehyung Kwon,
    Heechan Moon, 
    and Kijung Shin
    }
    \IEEEauthorblockA{\textit{Kim Jaechul Graduate School of AI,
    KAIST, Seoul, Republic of Korea}
    \\
    \{jhsk777, taehyung.kwon, heechan9801, kijungs\}@kaist.ac.kr}
    }

	\newcommand\red[1]{\textcolor{red}{#1}}
\newcommand\teal[1]{\textcolor{teal}{#1}}
\newcommand\blue[1]{\textcolor{blue}{#1}}
\newcommand\gray[1]{\textcolor{gray}{#1}}
\newcommand\green[1]{\textcolor{green}{#1}}

\definecolor{peace}{RGB}{55, 126, 184}
\definecolor{love}{RGB}{55, 126, 184}
\definecolor{joy}{RGB}{77, 175, 74}
\definecolor{kindness}{RGB}{152, 78, 163}

\newcommand\peace[1]{\textcolor{peace}{#1}}
\newcommand\love[1]{\textcolor{love}{#1}}
\newcommand\joy[1]{\textcolor{joy}{#1}}
\newcommand\kindness[1]{\textcolor{kindness}{#1}}

\newcommand\kijung[1]{\textcolor{peace}{[Kijung: #1]}}
\newcommand\fanchen[1]{\textcolor{love}{[Fanchen: #1]}}
\newcommand\minyoung[1]{\textcolor{joy}{[Minyoung: #1]}}
\newcommand\sunwoo[1]{\textcolor{kindness}{[Sunwoo: #1]}}
\newcommand{\taehyung}{\textcolor{magenta}}
\newcommand{\std}{\scriptsize}

\newcommand{\change}{\textcolor{red}}
\newcommand{\done}{\red{[DONE]}\xspace}
\newcommand{\todo}[1]{\red{[TODO: #1]}\xspace}
\newcommand{\ongoing}{\red{(ONGOING)}\xspace}
\newcommand{\tb}[1]{\textbf{#1}\xspace}
\newcommand{\repo}{\url{https://github.com/jhsk777/SPLASH}}

\newcommand{\smallsection}[1]{{\vspace{0.02in} \noindent {{\myuline{\smash{\bf #1:}}}}}}
\newtheorem{obs}{\textbf{Observation}}
\newtheorem{defn}{\textbf{Definition}}
\newtheorem{thm}{\textbf{Theorem}}
\newtheorem{axm}{\textbf{Axiom}}
\newtheorem{lma}{\textbf{Lemma}}
\newtheorem{cor}{\textbf{Corollary}}
\newtheorem{problem}{\textbf{Problem}}
\newtheorem{pro}{\textbf{Problem}}
\newtheorem{remark}{\textbf{Remark}}
\newtheorem{simpleexample}{\textit{Example}}
\newtheorem{note}{\textit{Note}}

\newcommand\und[1]{\underline{#1}}

\newcommand{\calA}{\mathcal{A}}
\newcommand{\calB}{\mathcal{B}}
\newcommand{\calC}{\mathcal{C}}
\newcommand{\calD}{\mathcal{D}}
\newcommand{\calE}{\mathcal{E}}
\newcommand{\calF}{\mathcal{F}}
\newcommand{\calG}{\mathcal{G}}
\newcommand{\calH}{\mathcal{H}}
\newcommand{\calI}{\mathcal{I}}
\newcommand{\calJ}{\mathcal{J}}
\newcommand{\calK}{\mathcal{K}}
\newcommand{\calL}{\mathcal{L}}
\newcommand{\calM}{\mathcal{M}}
\newcommand{\calN}{\mathcal{N}}
\newcommand{\calO}{\mathcal{O}}
\newcommand{\calP}{\mathcal{P}}
\newcommand{\calQ}{\mathcal{Q}}
\newcommand{\calR}{\mathcal{R}}
\newcommand{\calS}{\mathcal{S}}
\newcommand{\calT}{\mathcal{T}}
\newcommand{\calU}{\mathcal{U}}
\newcommand{\calV}{\mathcal{V}}
\newcommand{\calW}{\mathcal{W}}
\newcommand{\calX}{\mathcal{X}}
\newcommand{\calY}{\mathcal{Y}}
\newcommand{\calZ}{\mathcal{Z}}
\newcommand{\weightg}{\omega_{g}}
\newcommand{\weightm}{\omega_{m}}
\newcommand{\veca}{\boldsymbol{a}}
\newcommand{\vecb}{\boldsymbol{b}}
\newcommand{\vecc}{\boldsymbol{c}}
\newcommand{\vecd}{\boldsymbol{d}}
\newcommand{\vece}{\boldsymbol{e}}
\newcommand{\vecf}{\boldsymbol{f}}
\newcommand{\vecg}{\boldsymbol{g}}
\newcommand{\vech}{\boldsymbol{h}}
\newcommand{\veci}{\boldsymbol{i}}
\newcommand{\vecj}{\boldsymbol{j}}
\newcommand{\veck}{\boldsymbol{k}}
\newcommand{\vecl}{\boldsymbol{l}}
\newcommand{\vecm}{\boldsymbol{m}}
\newcommand{\vecn}{\boldsymbol{n}}
\newcommand{\veco}{\boldsymbol{o}}
\newcommand{\vecp}{\boldsymbol{p}}
\newcommand{\vecq}{\boldsymbol{q}}
\newcommand{\vecr}{\boldsymbol{r}}
\newcommand{\vecs}{\boldsymbol{s}}
\newcommand{\vect}{\boldsymbol{t}}
\newcommand{\vecu}{\boldsymbol{u}}
\newcommand{\vecv}{\boldsymbol{v}}
\newcommand{\vecw}{\boldsymbol{w}}
\newcommand{\vecy}{\boldsymbol{y}}
\newcommand{\vecz}{\boldsymbol{z}}

\newcommand{\vecsp}{\boldsymbol{s}'}

\newcommand{\matA}{\mathbf{A}}
\newcommand{\matB}{\mathbf{B}}
\newcommand{\matC}{\mathbf{C}}
\newcommand{\matD}{\mathbf{D}}
\newcommand{\matE}{\mathbf{E}}
\newcommand{\matF}{\mathbf{F}}
\newcommand{\matG}{\mathbf{G}}
\newcommand{\matH}{\mathbf{H}}
\newcommand{\matI}{\mathbf{I}}
\newcommand{\matJ}{\mathbf{J}}
\newcommand{\matK}{\mathbf{K}}
\newcommand{\matL}{\mathbf{L}}
\newcommand{\matM}{\mathbf{M}}
\newcommand{\matN}{\mathbf{N}}
\newcommand{\matO}{\mathbf{O}}
\newcommand{\matP}{\mathbf{P}}
\newcommand{\matQ}{\mathbf{Q}}
\newcommand{\matR}{\mathbf{R}}
\newcommand{\matS}{\mathbf{S}}
\newcommand{\matT}{\mathbf{T}}
\newcommand{\matU}{\mathbf{U}}
\newcommand{\matV}{\mathbf{V}}
\newcommand{\matW}{\mathbf{W}}
\newcommand{\matX}{\mathbf{X}}
\newcommand{\matY}{\mathbf{Y}}
\newcommand{\matZ}{\mathbf{Z}}

\newcommand{\method}{\textsc{SPLASH}\xspace}
\newcommand{\model}{\textsc{SLIM}\xspace}

\newcommand{\SedanSpot}{SedanSpot\xspace}
\newcommand{\MIDAS}{MIDAS\xspace}
\newcommand{\FFADE}{F-FADE\xspace}
\newcommand{\Anoedgel}{Anoedge-l\xspace}
\newcommand{\JODIE}{JODIE\xspace}
\newcommand{\Dyrep}{Dyrep\xspace}
\newcommand{\TGAT}{TGAT\xspace}
\newcommand{\TGN}{TGN\xspace}
\newcommand{\SAD}{SAD\xspace}
\newcommand{\MLP}{MLP\xspace}
\newcommand{\GCN}{GCN\xspace}
\newcommand{\GAT}{GAT\xspace}

\newcommand{\attention}{\textsc{ATT}\xspace}
\newcommand{\aggregation}{\textsc{AGG}\xspace}
\newcommand{\MultiheadAtt}{\textsc{MAB}\xspace}
\newcommand{\WithinAttentionFull}{within attention\xspace}
\newcommand{\WithinAttention}{\textsc{WithinATT}\xspace}
\newcommand{\PE}{\textsc{WithinOrderPE}\xspace}

\newcommand{\methodgrid}{\textsc{MiDaS-Grid}\xspace}
\newcommand{\methodauto}{\textsc{MiDaS}\xspace}

\definecolor{myred}{RGB}{195, 79, 82}
\definecolor{mygreen}{RGB}{86, 167 104}
\definecolor{myblue}{RGB}{74, 113 175}

\newcommand{\bigcell}[2]{\begin{tabular}{@{}#1@{}}#2\end{tabular}}

\let\oldnl\nl
\newcommand{\nonl}{\renewcommand{\nl}{\let\nl\oldnl}}

	\maketitle
        \begin{abstract}
		The problem of predicting node properties (e.g., node classes) in graphs has received significant attention due to its broad range of applications. Graphs from real-world datasets often evolve over time, with newly emerging edges and dynamically changing node properties, posing a significant challenge for this problem. 
In response, temporal graph neural networks (TGNNs) have been developed to predict dynamic node properties from a stream of emerging edges.
However, our analysis reveals that most TGNN-based methods are (a) far less effective without proper node features and, due to their complex model architectures, (b) vulnerable to distribution shifts.

In this paper, we propose SPLASH, 
a simple yet powerful method for predicting node properties on edge streams under distribution shifts.
Our key contributions are as follows: 
(1) we propose feature augmentation methods and an automatic feature selection method for edge streams, which improve the effectiveness of TGNNs,
(2) we propose a lightweight MLP-based TGNN architecture that is highly efficient and robust under distribution shifts, and
(3) we conduct extensive experiments to evaluate the accuracy, efficiency, generalization, and qualitative performance of the proposed method and its competitors on dynamic node classification, dynamic anomaly detection, and node affinity prediction tasks across seven real-world datasets.

	\end{abstract}
	\vspace{-1mm}
\section{Introduction}
\label{sec:intro}

Entities in many real-world networks have properties, and predicting these properties, which is naturally formulated as node property prediction on graphs, has been widely studied due to its importance in various applications~\cite{kipf2016semi,hamilton2017inductive, velickovic2017graph,gasteiger2018predict}. Notable examples include detecting fraud in financial networks~\cite{dou2020enhancing,liu2021pick}, predicting user interests in social networks~\cite{fan2019graph,guo2020deep}, and predicting users' purchasing affinity in e-commerce platforms~\cite{li2020hierarchical}.


Many real-world networks (e.g., social, financial, and purchase networks) evolve over time, with new edges emerging and node properties changing dynamically. As mentioned in ~\cite{tang2023dynamic,yang2024graphpro}, this evolution poses a critical challenge for node property prediction, as methods based on static graphs become less effective and efficient in such scenarios.


Graph stream algorithms~\cite{mcgregor2014graph,skarding2021foundations} are a class of algorithms designed to address this issue, and in them, time-evolving networks are modeled as streams of (emerging) edges, also known as continuous-time dynamic graphs (CTDGs).
These algorithms maintain 
intermediate results
and update them incrementally, as new edges arrive, to offer requested information in a timely manner. Stream-based methods are advantageous in terms of speed and space efficiency compared to static graph-based methods, which often require full re-computation to handle dynamic changes. This makes them particularly useful for time-critical applications.


Among graph stream algorithms, temporal graph neural networks (TGNNs)~\cite{kumar2019predicting, sankar2020dysat, xu2020inductive, tgn_icml_grl2020}
are particularly relevant to node property prediction. In response to each arriving edge, they dynamically update node representations that capture complex temporal and structural patterns to be used to predict dynamic node properties. To achieve this, TGNNs utilize complex neural network architectures, often incorporating recurrent neural networks, self-attention mechanisms, and memory modules.




In this work, we focus on two aspects that have been overlooked in the existing TGNN methods: \textbf{node features} and \textbf{distribution shifts}.
Our analysis reveals that most TGNN methods are (a) significantly less effective without proper node features and are (b) vulnerable to distribution shifts.
A detailed discussion of these limitations is provided in Section~\ref{sec:related:findings}.

To address these limitations, we propose \method (\underline{\smash{S}}imple node \underline{\smash{P}}roperty prediction via representation \underline{\smash{L}}earning with \underline{\smash{A}}ugmented features under distribution \underline{\smash{SH}}ifts).
Guided by the above analysis, \method augments node features that encode positional and structural information from edge streams, significantly enhancing prediction performance.
Especially, \method automatically selects feature augmentation schemes based on empirical risks, without requiring any prior knowledge. 
Lastly, instead of complex architectures, \method employs a lightweight MLP-based model, resulting in an improved generalization capability under distributional shifts.


We consider various node property prediction tasks (spec., classification, dynamic anomaly detection, and
node affinity prediction) in our experiments using seven real-world datasets. The results reveal the following advantages of \method:
\begin{itemize}[leftmargin=*]
    \item \textbf{Fast \& lightweight}: \method uses only MLP layers, enabling fast inference.
    It is up to 27.52$\times$ faster with up to 5.97$\times$ fewer parameters than best-performing competitors.
    \item \textbf{Effective}: \method significantly and consistently outperforms all baselines, especially under distribution shifts, with prediction performance gains of up to 13.55\%.
    \item \textbf{Automatic}: \method accurately selects feature augmentation schemes, without external knowledge or tuning.
\end{itemize}
For \textbf{reproducibility}, we provide our code and datasets at \repo.



	\section{Preliminaries and Related Works}
\label{sec:related:others}
In this section, we cover preliminaries and related work.
Frequently used notations are summarized in Table~\ref{tab:notation_table}.

\subsection{Continuous Time Dynamic Graph}
\label{sec:related:ctdg}

\smallsection{Definition and Related Concepts}
A continuous-time dynamic graph (CTDG) $\mathcal{G}=(\delta^{(1)}, \delta^{(2)}, \cdots)$ is a continuous stream of temporal edges, each with an associated timestamp.
A temporal edge $\delta^{(n)}=(v_{i}, v_{j},\mathbf{x}_{ij}^{(n)},w_{ij}^{(n)}, t^{(n)})$, arriving at time $t^{(n)}\in \mathcal{I}$, is directed from the \textit{source node} $v_i$ to the \textit{destination node} $v_j$ with an edge feature $\mathbf{x}_{ij}^{(n)}\in \mathbb{R}^{d_e}$ and edge weight $w_{ij}^{(n)}\in \mathbb{R}$, where $\mathcal{I}$ denotes a set of possible timestamps and $d_e$ indicates a dimension of edge features. 
The temporal edges are ordered chronologically, i.e., $t^{(n)}\leq t^{(n+1)}$ holds for all $n \in \{1,2,\cdots\}$.
We denote the set of nodes in $\mathcal{G}$ as $\mathcal{V}=\bigcup_{\delta^{(n)} \in \mathcal{G}}\{v_{i},v_{j}\}$. 
In addition, we denote the graph snapshot accumulated up to time $t^{(n)}$ as $\bold{G}^{(n)}=(\mathcal{V}^{(n)}, \mathcal{E}^{(n)}, \Omega^{(n)})$, where $\mathcal{V}^{(n)}$ is the set of nodes, $\mathcal{E}^{(n)}$ is the set of edges, and $\Omega^{(n)}$ is the edge weight function.
Formally, 
$\mathcal{V}^{(0)}=\mathcal{E}^{(0)}=\emptyset$, $\Omega^{(0)}(\cdot)=0$, and for each $\delta^{(n)}=(v_{i}, v_{j},\mathbf{x}_{ij}^{(n)},w_{ij}^{(n)}, t^{(n)})$ with $n\geq 1$, the following equalities hold: 
    $\mathcal{V}^{(n)}  = \mathcal{V}^{(n-1)}\cup\left\{v_i,v_j \right\}$, 
     $\mathcal{E}^{(n)} =\mathcal{E}^{(n-1)}\cup\left\{(v_i,v_j) \right\}$,
     $\Omega^{(n)}((v_i,v_j)) =\Omega^{(n-1)}((v_i,v_j))+w_{ij}^{(n)}$, and
     $\Omega^{(n)}(e) =\Omega^{(n-1)}(e), \forall e \in \mathcal{E}^{(n)} \setminus \{(v_i,v_j)\}$.

\smallsection{Advantages}
A CTDG is a natural data structure for representing time-evolving networks, and it is well-suited for real-time processing where a small batch of emerging edges or even each individual edge serves as processing units.
Additionally, most algorithms on CTDGs can be memory-efficient for large-scale networks, typically storing only smaller intermediate results rather than all past edges.


\smallsection{Applications}
CTDGs have been applied to time-critical tasks on evolving networks, including
anomaly detection~\cite{eswaran2018sedanspot, bhatia2020midas,chang2021f}, user classification~\cite{zhou2022tgl}, and item recommendation~\cite{kumar2019predicting, severin2023ti, huang2024temporal}, where capturing up-to-date information as soon as it arrives and providing timely responses are crucial.



\subsection{Temporal Graph Neural Networks (TGNNs)}
\smallsection{Overview}
Temporal graph neural network (TGNN) is a class of neural networks designed for representation learning on CTDGs.
TGNNs generally update node representations whenever a new edge arrives by message passing between neighboring nodes, which is a common technique in GNNs. 
Typically, edges up to a specific time point are used to train TGNN parameters, and then the trained parameters are applied to update node representations based on subsequent edges. The updated node representations are used for downstream tasks.

\begin{table}[t]
    \centering
    \caption{List of frequently used notations.}
    \scalebox{0.9}{
        \begin{tabular}{c|l}
            \toprule
              \textbf{Notations} & \textbf{Descriptions}  \\
            \midrule
            $\mathcal{G}$  & Input continuous-time dynamic graph (CTDG)\\
            $\mathcal{I}$ & Set of possible timestamps \\
            $t_{seen}$ & End time of the training period \\
            $\mathcal{G}_{<t}, \mathcal{G}_{seen}$ & $\mathcal{G}$ up to time $t$ and $\mathcal{G}$ up to time $t_{seen}$ for training \\ 
            $\mathcal{V}$, $\mathcal{V}_{seen}$ & Node sets in $\mathcal{G}$ and  $\mathcal{G}_{seen}$ \\ 
            $v_i$ & Node with index $i$ \\ 
            $\delta^{(n)}$ &  Temporal edge of order $n$ in $\mathcal{G}$ \\
            $\textbf{x}_{ij}^{(n)}$, $w_{ij}^{(n)}$ & Edge feature and weight of $\delta^{(n)}$ between $v_i$ and $v_j$\\             
            $\bold{G}^{(n)}$ & Graph snapshot accumulated from $\mathcal{G}$  up to time of $\delta^{(n)}$ \\
            $\mathcal{V}^{(n)}$, $\mathcal{E}^{(n)}$ &  Node and edge sets in $\bold{G}^{(n)}$ \\
            $\Omega^{(n)}$ & Edge weight function for $\bold{G}^{(n)}$ \\
            $d_e$, $d_v$  & Dimensions of edge features and node features  \\          
            \midrule
            $R$, $P$, $S$  & Random, positional, and structural feature augmentation processes  \\
            \midrule
            $\boldsymbol{r}_i(t)$, $\boldsymbol{p}_i(t)$, & Augmented random, positional, and structural features of $v_i$  \\
            $\boldsymbol{s}_i(t)$ &  at time $t$  \\            
                 \midrule
            $X$  & General node feature augmentation process, i.e., $X\in \left\{R,P,S \right\}$  \\
            $\boldsymbol{x}_i(t)$  & Node feature of $v_i$ at time $t$, i.e., $\boldsymbol{x}_i(t)\in\left\{\boldsymbol{r}_i(t),\boldsymbol{p}_i(t),\boldsymbol{s}_i(t) \right\}$  \\
            $X^{\star}$  & Selected node feature augmentation process \\

            $\boldsymbol{x}_i^{\star}(t)$  & Selected node feature of $v_i$ at time $t$ generated from $X^{\star}$  \\
            $\mathcal{N}_{i}(t)$  & Set of the most k recent temporal edges of $v_i$ at time $t$ \\
            $Y_{i}(t)$  & Property label of $v_i$ at time $t$ \\
            \bottomrule
        \end{tabular}
    }
    \label{tab:notation_table}
\end{table}

\smallsection{Example TGNNs}
JODIE~\cite{kumar2019predicting} utilizes recurrent neural network~\cite{rumelhart1986learning} modules to update dynamic node representations (i.e., node representations that evolve over time) by sequentially encoding interaction histories of nodes.
DySAT~\cite{sankar2020dysat}, especially its CTDG variant  \cite{zhou2022tgl}, converts a CTDG into graph snapshots to apply GAT~\cite{velickovic2017graph} to each snapshot, 
and it uses the self-attention mechanism of transformers~\cite{vaswani2017attention} along the temporal dimension.
TGAT~\cite{xu2020inductive} leverages temporal encoding and graph attention
to incorporate temporal information in generating dynamic node representations.
TGN~\cite{tgn_icml_grl2020} employs a memory module to capture the long-term temporal and structural interaction patterns of each node. 
GraphMixer~\cite{cong2022we} aims to generate edge representations utilizing MLP-mixer~\cite{tolstikhin2021mlp} architectures, focusing on edge features and time information.
DyGFormer~\cite{yu2023towards} captures correlations of node pairs within edges using neighbor co-occurrence information of the node pairs as encodings for transformers to capture long-term temporal dependencies more effectively.


\smallsection{Advantages}
The primary advantage of TGNNs is their ability to apply message passing incrementally to CTDGs for time-critical tasks.
Note that message passing between neighboring nodes is a key technique of GNNs for effectively modeling complex relationships in graphs. 
TGNNs often better address complex tasks that rule-based approaches have struggled to solve in CTDGs.
During the inference stage,
TGNNs typically process each arriving edge in constant time, regardless of the overall size of the CTDG.

\smallsection{Applications}
TGNNs have been employed to address complex tasks in CTDGs, including anomaly detection~\cite{tian2023sad,lee2024slade}, node classification~\cite{zhou2022tgl,xu2024scalable}, node affinity prediction~\cite{huang2024temporal},
link prediction~\cite{cong2022we,yu2023towards}, and recommendation~\cite{kumar2019predicting, zhao2023time}.
Typically, TGNNs~\cite{kumar2019predicting, sankar2020dysat, xu2020inductive, tgn_icml_grl2020} are used to produce dynamic node representations, and they are fed into classifiers~\cite{zhou2022tgl, xu2024scalable}, regressors~\cite{huang2024temporal}, and anomaly detectors~\cite{tian2023sad, lee2024slade}.
Notably, TGNNs are often trained without label supervision when applied to anomaly detection~\cite{tian2023sad, lee2024slade}.

\begin{figure}[!t]
    \centering
    \vspace{-2mm}
    \includegraphics[width=0.48\textwidth]
    {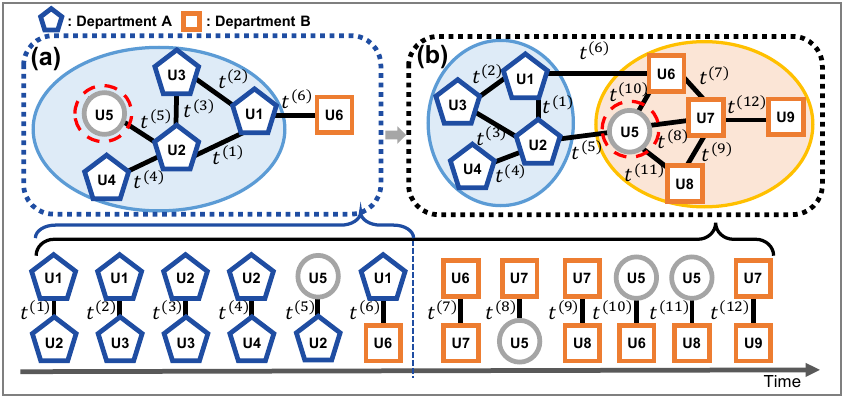}
    \vspace{-2mm}
    \caption{
        An example of distribution shifts in a collaboration network from a company with two departments. (a) shows the network before the distribution shift at time $t^{(6)}$, and (b) shows the network after the shift. 
        Note that node U5's community membership shifts from Department A to B over time.
    }
    \label{fig:dist_shift_simple_example}
\end{figure}

\subsection{Distribution Shifts}
\label{related:ds}

\smallsection{Definition}
A distribution shift refers to a change in the underlying data distribution between training and test sets.
Such shifts degrade the generalization ability of trained models, resulting in poor performance on unseen data that differs from the training distribution.
Distribution shifts can occur due to various factors, including temporal changes~\cite{koh2021wilds}, domain changes~\cite{tachet2020domain}, and sample bias~\cite{liu2014robust}.
Distribution shifts have been studied across multiple domains, including natural language processing~\cite{yuan2023revisiting,hendrycks2020pretrained} and computer vision~\cite{shu2022test,koh2021wilds}, as well as graph learning~\cite{wuhandling,yoo2023disentangling}.

\smallsection{Distribution Shifts on Temporal Networks}
 In temporal networks, distribution shifts due to temporal changes can arise in various forms, including shifts in (a) positional distributions (e.g., community memberships of nodes), (b) structural distributions (e.g., node degrees), and (c) property distributions (e.g., external node labels or features) over time.
 In addition to Example~\ref{example:shift}, refer to Section~\ref{sec:related:findings} (Figure~\ref{fig:shift_example_intro}) for example distribution shifts in real-world temporal networks.

\begin{simpleexample}\label{example:shift}
    Figure~\ref{fig:dist_shift_simple_example} shows an example where a distribution shift occurs at time $t^{(6)}$ in a collaboration network from a company with two departments.
    Before $t^{(6)}$, node U5 forms a community with nodes from Department A.
After $t^{(6)}$, U5 forms a new community with nodes from Department B, demonstrating a distribution shift over time.
\end{simpleexample}

 A common approach for addressing such distribution shifts in temporal network methods~\cite{zhang2022dynamic,yuan2024environment,zhang2024spectral} is to generate multiple representations through disentangled representation learning. This approach separates the underlying factors of variation within the data to create distinct representations that capture patterns invariant under distribution shifts.
 However, this approach requires access to the entire graph as a whole, making it difficult to apply to CTDGs.


\smallsection{(Remaining Challenge) Distribution Shifts in TGNNs}
TGNNs are typically trained on past edges up to a specific time point and tested on consecutive future edges. Thus, the aforementioned temporal changes in the underlying networks lead to distribution shifts between the training and test sets.

TGNNs are especially vulnerable to such distribution shifts because their high complexity makes them prone to overfitting the training set, resulting in a loss of generalization ability, as empirically confirmed in Section~\ref{sec:exp:experiment_results} (see Figure~\ref{fig:generalization_ratio}).

Furthermore, many techniques for addressing distributional shifts in general GNNs, including the ones mentioned earlier, are inapplicable to TGNNs due to their difference. 
In the test (i.e., inference) stage,
GNNs typically require access to the entire graph to construct node representations. In contrast, TGNNs are designed for CTDGs, and thus, node representations are incrementally updated based on each batch of edges (or even individual edges), without access to the entire graph.

\subsection{Node Feature Augmentation}
\label{subsec:node_feat_aug}

\smallsection{Overview}
Node features are numerical or categorical attributes associated with each node in a graph. 
Combined with the graph structure, they are often used as input to GNNs for various tasks \cite{duong2019node,cui2022positional}.
Node features can be externally provided or, as described below, augmented.
External node features offer information beyond the graph structure.

\vspace{-0.1cm}
\begin{simpleexample}
In citation networks between papers, the abstract of a paper can be converted into a vector using bag-of-words representations to be used as an external node feature.
\end{simpleexample}

\begin{figure}[!t]
\vspace{-2mm}
    \centering
    \includegraphics[width=0.48\textwidth]
    {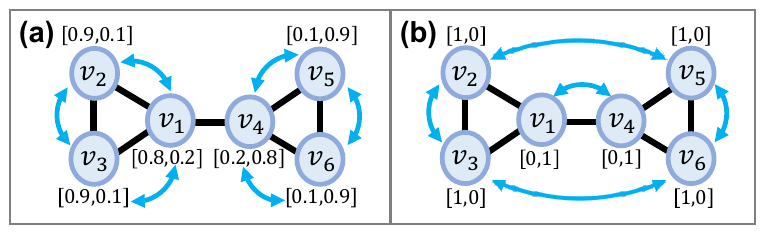}
    \vspace{-2mm}
    \caption{
        An example graph with (a) positional node features and (b) structural node features.
        The blue arrows indicate node pairs with similar node features.
    }
    \label{fig:augment_example}
\end{figure}

\smallsection{Definition of Node Feature Augmentation}
\label{subsec:node_feat_aug:def}
Node feature augmentation refers to generating artificial node features.
These augmented features can be positional or structural, capturing nodes' positional or structural properties in a graph.
Augmentation enriches node information, especially when external node features are missing or weakly informative. 


\begin{simpleexample}
    As shown in Figure~\ref{fig:augment_example}(a), 
    positional node features are similar among spatially close neighbors, while as shown in Figure~\ref{fig:augment_example}(b), structural node features are similar among nodes with similar structural characteristics, such as degree.
\end{simpleexample}



A wide range of node embedding techniques can be employed for this purpose. Given their variety, we refer readers to surveys~\cite{zhou2022network,cui2018survey,barros2021survey} for a comprehensive overview.
Below, we briefly introduce a few representative ones, categorized into positional embeddings for positional node features and structural embeddings for structural node features.

\smallsection{Positional Embeddings}
Positional node embedding aims to capture the positions of nodes within a graph by assigning similar features to spatially close nodes, such as those within few hops (e.g., GraRep \cite{cao2015grarep}) and those frequently co-occurring in random walks (e.g., DeepWalk  \cite{perozzi2014deepwalk} and node2vec \cite{grover2016node2vec}).

\smallsection{Structural Embeddings}
Structural node embedding aims to capture the structural characteristics of nodes by assigning similar features to nodes with analogous structural properties. Examples include one-hot vectors based on node degree~\cite{houmeasuring,hamilton2017inductive}, PageRank scores~\cite{brin1998anatomy}, and embeddings that incorporate the structural properties of not just each node but also its (multi-hop) neighbors (e.g., struc2vec~\cite{ribeiro2017struc2vec}).

\smallsection{(Remaining Challenge) Feature Augmentation for TGNNs}
Despite this abundance of effective embedding methods, they have yet to be combined with TGNNs for input feature augmentation. For featureless graphs,  existing TGNNs~\cite{kumar2019predicting, xu2020inductive, tgn_icml_grl2020} have ignored node features or used zero vectors.
This is likely because existing embedding methods are not directly applicable to CTDGs or TGNNs, given (a) the limited access to the input graph, (b) the requirement for real-time processing, and (c) the challenge of handling new nodes unseen during training.
To be applied to TGNNs, embedding methods need to be adapted to generate embeddings for unseen nodes rapidly, using only limited historical data.

\subsection{Graph Stream Processing}
\smallsection{Overview}
Graph stream processing refers to incrementally solving tasks as new edges and queries arrive over time, with a focus on reducing memory usage and running time.




\smallsection{Tasks and Efficiency of Graph Stream Processing}
Typical graph streaming algorithms focus on computational tasks, such as computing connectivity~\cite{feigenbaum2005graph}, generating cut sparsifier~\cite{ahn2009graph}, finding densest subgraphs~\cite{bhattacharya2015space}, and counting triangles~\cite{pavan2013counting,wang2017approximately,lee2020temporal}, by directly computing or approximating the target quantity or structure.
In addition to the exact storage of graph streams~\cite{sheng2018grapu,mariappan2021dzig,feng2021risgraph,feng2015distinger, sengupta2016graphin,macko2015llama, mariappan2019graphbolt},
approximation and summarization techniques for graph streams~\cite{ahn2012graph,kapralov2014spanners,kallaugher2018sketching,kane2012counting,ko2020incremental,tang2016graph,zhao2011gsketch} have been explored to efficiently preserve general or task-specific information. 
As noted in \cite{mcgregor2014graph} and \cite{besta2021practice}, by summarizing the input graph stream, graph stream algorithms typically achieve space requirements sub-linear to the total number of edges.

\smallsection{Connection to Our Study}
Since our approach targets machine learning tasks rather than computational tasks, its mechanism based on node representation learning differs fundamentally from conventional graph stream algorithms. Despite the difference, the high-level goal remains the same: to solve given tasks rapidly and efficiently. To this end, instead of storing the entire graph, we maintain a summary that limits the number of neighbors per node, 
ensuring sub-linear space requirements in terms of the total edge count.
Regarding efficiency, also refer to the time complexity in Section~\ref{sec:method} and the empirical results in Section~\ref{sec:exp:efficiency}.

\vspace{-2mm}



\begin{figure}[!t]
    \centering
    \vspace{-2mm}\includegraphics[width=0.48\textwidth]
    {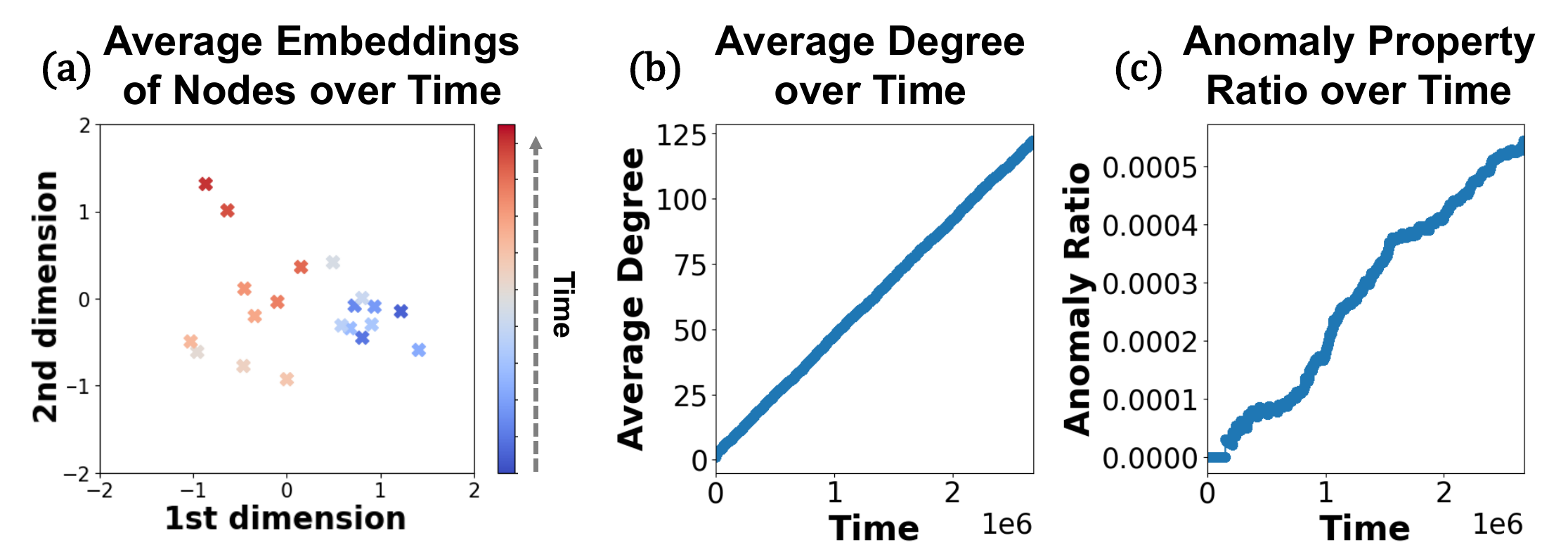}
    \vspace{-2mm}
    \caption{
        Examples of distribution shifts in edge streams: (a) positional, (b) structural, and (c) property distribution shifts over time in the Reddit dataset. 
        In (a), nodes are grouped based on their appearance time, and the node embeddings generated by node2vec~\cite{grover2016node2vec} using the entire graph are averaged within each group. 
        These averaged embeddings are visualized using t-SNE.
    }
    \label{fig:shift_example_intro}
\end{figure}

\subsection{Preliminary Analysis on the Limitations of TGNNs}
\label{sec:related:findings}
Below, we provide a summary of our findings from the preliminary analysis on the limitations of TGNNs.

\smallsection{Our Findings regarding Node Features}
Most TGNNs can leverage node features as input; however, many real-world time-evolving graphs lack node features entirely or have only weakly informative features.
In such cases, it is common to omit node features or use zero vectors as input, but this often results in a significant drop in node property prediction performance.
Interestingly, even a simple augmentation of node features (spec., assigning distinct randomly generated features to each node) can lead to substantial performance improvements.
Detailed experimental evidence (e.g., dynamic node classification results in Table~\ref{tab:main_performance}) is provided in Section~\ref{sec:exp:experiment_results}. 

\smallsection{Our Findings regarding Distribution Shifts}
\label{intro:finding_dist}
Distribution shifts over time are commonly observed in real-world graphs~\cite{zhang2022dynamic, zhang2024spectral, ko2022begin}. 
Examples include (a) positional and (b) structural shifts, which are caused by newly emerging edges, and (c) node-property distribution shifts, as illustrated in Figure~\ref{fig:shift_example_intro}.
Under distribution shifts, TGNNs, which are typically trained on past data and make inferences on future data, easily struggle with generalization due to their complex architectures, yielding inaccurate node-property predictions.
Refer to Section~\ref{sec:exp:experiment_results} for empirical evidence.

\section{Problem Description}
\label{sec:prelim}

In this section, we introduce our target task, node property prediction in continuous-time dynamic graphs (CTDGs).


\begin{figure}[!t]
    \centering
    \vspace{-2mm}\includegraphics[width=0.48\textwidth]
    {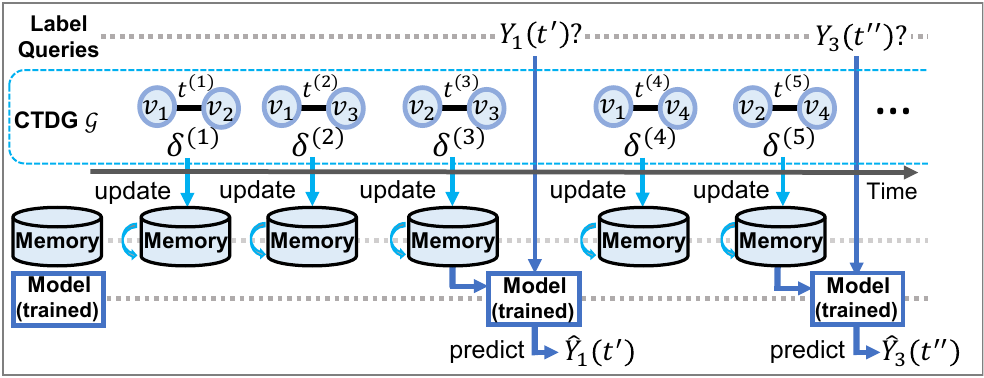}
    \vspace{-2mm}
    \caption{
        An example of node property prediction in a CTDG over time. This process involves a memory that stores a summary or sample of the CTDG.
        Whenever a temporal edge arrives, the memory is updated, and if a label query is received, a model (e.g., TGNN) makes a prediction based on the memory updated until that time point.
    }
    \label{fig:CTDG_example}
\end{figure}

\smallsection{Problem Definition}
The node property prediction task on an evolving network involves predicting the property of each node at each time point $t$, using the temporal edges up to $t \in \mathcal{I}$.
Due to the advantages discussed in Section~\ref{sec:related:ctdg}, 
the evolving network is modeled as a CTDG, and we define the task using the notations defined in Section~\ref{sec:related:ctdg}.
Given a CTDG $\mathcal{G}=(\delta^{(1)}, \delta^{(2)}, \cdots$), our goal at each time $t \in \mathcal{I}$ is to accurately predict the label $Y_i(t)$ for every node
$v_{i}$ that has appeared up to $t$.
Note that 
predictions are based only on the edges that have arrived up to time $t$ (i.e., $\{\delta^{(l)} \in \mathcal{G} : t^{(l)}\leq t\}$), and
future edges in $\mathcal{G}$ (i.e., $\{\delta^{(l)} \in \mathcal{G} : t^{(l)}>t\}$) are not accessible at $t$.
\begin{simpleexample}
Figure~\ref{fig:CTDG_example} shows an example of node property prediction in a CTDG where temporal edges and label queries arrive over time.
Whenever a new temporal edge arrives, the memory storing a summary or sample of the CTDG is updated.
Given a label query, a model (e.g., TGNN) makes a prediction based on the memory updated until that time point.
\end{simpleexample}

The form of the labels, which we aim to predict, varies 
across task instances, as described below. 



\smallsection{Example 1 - Dynamic Node Classification}
In dynamic node classification on CTDGs, we aim to predict the class of each node at each time, i.e., $Y_i(t)\in C$, where $C$ is the set of classes.
We specifically consider a semi-supervised setting where labels are available only for a subset of nodes seen during training, while the labels for other seen nodes and unseen nodes appearing after training remain unknown.
Note that, unlike node classification on static graphs, the class of a node may change over time~\cite{kumar2019predicting, lee2024slade}.

\smallsection{Example 2 - Dynamic Anomaly Detection}
In dynamic anomaly detection on CTDGs, the state of each node at each time,
which can be either normal or abnormal,
is treated as the property that we aim to predict, i.e., $Y_i(t)\in\left\{\textit{normal},\textit{abnormal} \right\}$.
Technically, this is a special case of dynamic node classification, but we treat it as a separate task due to the existence of approaches dedicated to anomaly detection, which leverage behavioral cues in addition to or instead of label supervision \cite{eswaran2018sedanspot,bhatia2020midas,chang2021f,lee2024slade}.

\smallsection{Example 3 - Node Affinity Prediction}
In node affinity prediction \cite{huang2024temporal} on CTDGs, the future affinities of each node to all or a subset of other nodes are treated as the property that we aim to predict, i.e., $Y_i(t)\in\mathbb{R}^{d_a}$, where $d_a$ is the number of nodes with which affinity is possible.
Specifically, affinities at each time point are given as the weights of temporal edges in the input CTDG, and at each time $t$, we aim to predict the normalized sum of affinities over the future period $\left [t, t+T_w  \right ]$, where $T_w$ is application-dependent (e.g., a week and a year).
Predicting time-evolving affinities can be valuable for various applications, including recommendations~\cite{tang2023dynamic,huang2024temporal}.

	\vspace{-1mm}

\section{Proposed Method: \method}
\label{sec:method}

\begin{figure}[!t]
    \centering
    \includegraphics[width=0.48\textwidth]
    {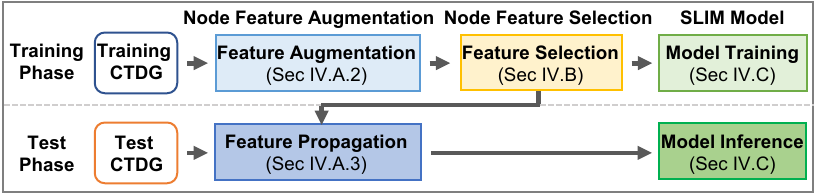}
    \caption{
        An outline of \method. In the training phase, for a given training CTDG, \method (1) generates augmented node features through feature augmentation, (2) identifies task-relevant features using feature selection, and (3) trains our proposed SLIM model with the selected augmented features.
        In the test phase, for a given test CTDG, \method (1) generates the selected augmented features for nodes unseen during training through feature propagation and (2) predicts node properties using the trained SLIM model.
    }
    \label{fig:overall_method}
\end{figure}

\begin{figure*}[t]
    \centering
    \vspace{-2mm}
    \includegraphics[width=1\linewidth]{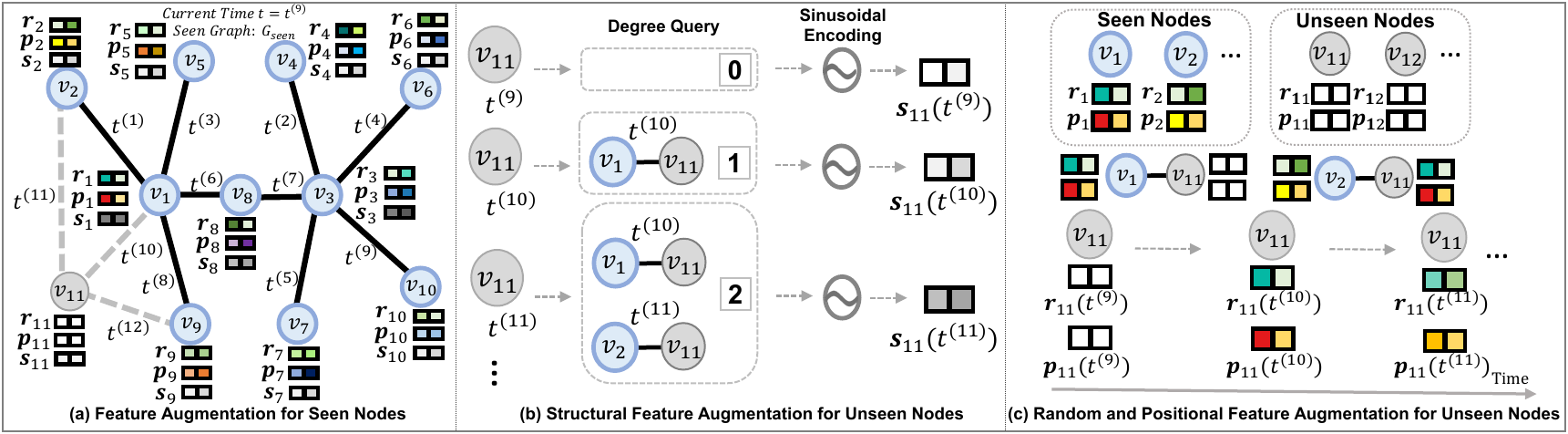}
    \centering
    \vspace{-6mm}
    \caption{ \label{fig:feature_augmentation}
        Overview of node feature augmentation in \method.
        (a) First, \method encodes the positional or structural characteristics 
        of seen nodes in the training period
        to generate their node features.
        Subsequently, for (b) structural node feature augmentation, \method assigns node features to unseen nodes by encoding their degrees, which are incrementally computed, while
        for (c) positional and random node feature augmentation, \method incrementally updates the features of unseen nodes by propagating the features of seen nodes.
    }
\end{figure*}

In this section, we present \method (\underline{\smash{S}}imple node \underline{\smash{P}}roperty prediction via representation \underline{\smash{L}}earning with \underline{\smash{A}}ugmented features under distribution \underline{\smash{SH}}ifts), our proposed method for node property prediction in CTDGs.
\method employs feature augmentation and a novel lightweight TGNN to enhance effectiveness, especially under distributional shifts.
Specifically, given a CTDG,
\method first augments node features, 
as we describe in Section~\ref{sec:method:augmentation}. 
It then performs automatic feature selection, as mentioned in Section~\ref{sec:method:selection}.
Lastly,
\method employs a novel lightweight MLP-based TGNN to predict the node property, based on the selected augmented node features and the CTDG, as we describe in Section~\ref{sec:method:model}.
Figure~\ref{fig:overall_method} provides an outline of \method, illustrating how its components are composed for the training and testing phases.
\vspace{-2mm}

\subsection{Node Feature Augmentation}
\label{sec:method:augmentation}



In this subsection, we describe our approach to augmenting node features in a CTDG. In general, generating node features and incorporating them as additional inputs to GNNs can enhance the informativeness of node representations.
However, as discussed in Section \ref{subsec:node_feat_aug},  applying existing node feature augmentation methods to CTDGs poses significant challenges: (a) the limited access to the input graph, (b) the need for real-time processing, and (c) the emergence of new nodes unseen during training.
Consequently, feature augmentation for TGNNs specialized to CTDGs remains unexplored, despite its effectiveness demonstrated in Section~\ref{sec:exp:experiment_results} (refer to Table~\ref{tab:main_performance}).

\begin{note}[Relation with Other Components]
Node feature augmentation generates multiple augmented node features.
Among them, \textbf{node feature selection} (Section~\ref{sec:method:selection}) selects the most effective augmented node features for our scenario. 
Then, the \textbf{\model model} (Section~\ref{sec:method:model})  utilizes the selected augmented node features as input for training and inference.
\end{note}


\subsubsection{Overview}
We propose a node augmentation method that extends existing node embedding techniques for their application to CTDGs, addressing the aforementioned challenges.
Our method has two steps:
(a) generating features for nodes that appear within the training period
and (b) generating features for nodes that appear after the training period, using feature propagation.
In this context, feature propagation refers to a process of spreading features across a graph, particularly from nodes with existing (augmented) features to nodes without features.
Feature propagation is performed incrementally, without incurring a significant computational cost, making it suitable for CTDGs.
Our method generates random, positional, and structural features, which are created by three different feature augmentation processes.
For each process $X$, we use $\bold{x}_i(t)=X(v_i(t))$ to denote the feature vector $\bold{x}_i\in\mathbb{R}^{d_v}$  for node $v_i$ at time $t$ with a node feature dimension $d_v$.
A visual overview of the proposed node feature augmentation method is given in Figure~\ref{fig:feature_augmentation}, and each step is described in detail below.

\subsubsection{Feature Augmentation on Training Graphs} 
As the first step, we generate features for nodes that appeared within the training period, referred to as \textit{seen nodes}.
If the training period of the input CTDG ends at time $t_{seen}$, the set of temporal edges arriving before $t_{seen}$ is used for training, and it is denoted as $\mathcal{G}_{seen} =(\delta^{(1)}, \delta^{(2)}, \cdots, \delta^{(s)})$, where $t^{(s)}\leq t_{seen}<t^{(s+1)}$.
Note that
$t^{(s)}$ denotes the largest timestamp in $\mathcal{G}_{seen}$, and $\mathcal{V}_{seen}=\mathcal{V}^{(s)}$ denotes the set of seen nodes.
For example, in Figure~\ref{fig:feature_augmentation}(a), $G_{seen}$ is $(\delta^{(1)},\delta^{(2)},\dots,\delta^{(9)})$; $V_{seen}$ is $\{v_1,v_2,v_3,v_4,v_5,v_6,v_7,v_8,v_9,v_{10}\}$; and $t_{seen}$ is $t^{(9)}$.
As is common in TGNN literature, we assume that the edges within the training period are few enough to be fully maintained, and thus the graph snapshot $\bold{G}^{(l)}=(\mathcal{V}^{(l)}, \mathcal{E}^{(l)}, \Omega^{(l)})$ at a time $t^{(l)}$ (see Section~\ref{sec:related:ctdg} for its definition) is available.
Features for seen nodes can be obtained by applying any existing node embedding techniques to the snapshot $\bold{G}^{(l)}$.
In this work, we employ three simple embedding processes that are both faster and empirically effective, especially under distribution shifts.

\smallsection{Process 1 - Random Feature Augmentation}
This process aims to encode the stable and absolute positions of seen nodes by simply assigning random vectors for seen nodes drawn from Gaussian Distribution for each dimension as $\bold{r}_{i}\sim \mathcal{N}(\bold{0},\bold{I})$, where $\bold{r}_{i}$ is a random feature vector of $v_{i}\in\mathcal{V}_{seen}$.
In this case, random feature vectors of seen nodes are fixed over time, i.e., $\bold{r}_i(t)=\bold{r}_i, ^{\forall}{v_i\in\mathcal{V}_{seen}}$ at any given time $t$, representing their temporally stable and absolute positions in high-dimensional space.
We denote the random feature augmentation process as $R$, i.e., $\bold{r}_i(t)= R(v_i(t))$.
\begin{simpleexample}
    As shown in Figure~\ref{fig:feature_augmentation}(a), random features (in shades of green) are assigned without any structural or positional pattern and serve solely to distinguish node identities. 
\end{simpleexample}

\smallsection{Process 2 - Positional Feature Augmentation}
This process aims to address the limitation of random features in capturing the proximity between nodes in the graph.
We simply apply a positional embedding method, which is based on proximity between nodes, to the training graph snapshot $\bold{G}^{(s)}$, as follows:
\vspace{-0.23cm}
\begin{equation}\label{eq:transductive1}
\vspace{-0.23cm}
\small
\bold{p}_{i}=Embedding( \bold{G}^{(s)}(v_i, \mathcal{V}^{(s)}, \mathcal{E}^{(s)}, \Omega^{(s)})),
\end{equation}
where $Embedding$ is a positional embedding function (see Section~\ref{subsec:node_feat_aug} for examples) that outputs a feature vector for a given node and in a given graph; in this work, we use node2vec \cite{grover2016node2vec} as the function.
Note that $\bold{p}_i$ is a positional node feature vector of $v_i\in\mathcal{V}_{seen}$.
Similar to the random feature augmentation, positional feature vectors of seen nodes $v_i\in\mathcal{V}_{seen}$ are fixed over time, i.e., $\bold{p}_i(t)=\bold{p}_i, ^{\forall}{v_i\in\mathcal{V}_{seen}}$ at any given time $t$, representing their temporally stable and relative positions in $\mathcal{G}_{seen}$.
We denote the positional feature augmentation process as $P$, i.e., $\bold{p}_i(t)= P(v_i(t))$.
\begin{simpleexample}
As shown in Figure~\ref{fig:feature_augmentation}(a), positional features are generated to assign similar features to nodes that are locally nearby in $\mathcal{G}_{seen}$. Note that positional features for the nodes $v_{1}$, $v_{2}$, $v_{5}$, and $v_{9}$ are in shades of red, and those for nodes $v_{3}$, $v_{4}$, $v_{6}$, $v_{7}$, and $v_{10}$ are in shades of blue.
\end{simpleexample}

\smallsection{Process 3 - Structural Feature Augmentation}
This process aims to encode the dynamic structural patterns of seen nodes. To this end, it leverages node degrees, which are basic structural characteristics.
The degree of the seen node $v_i\in\mathcal{V}_{seen}$ at a specific time $t$ can be defined as follows:
\begin{equation}\label{eq:transductive1}
\small
deg_i(t)=\sum\nolimits_{(v_{i}, v_{j},\bold{x}_{ij}^{(n)},w_{ij}^{(n)}, t^{(n)}) \in \mathcal{G}}\mathbb{I}(t^{(n)} \leq t).
\end{equation}
Thus, the degree of each node can be incrementally updated whenever a new temporal edge involving the node appears.
Instead of one-hot encoding, which requires varying lengths as the maximum degree changes over time, we generate a structural node feature vector of $v_i\in\mathcal{V}_{seen}$ at time $t$ by encoding its corresponding degree using sinusoidal encoding: 
\begin{equation}
\small
\left [\bold{s}_i(t) \right ]_{n} = \left [\phi_{d}(deg_i(t)) \right ]_{n} =
\begin{cases}
\cos\left(\alpha^{-\frac{n}{2\sqrt{d_v}}} deg_i(t)\right), & \text{if }n \text{ is even} \\
\sin\left(\alpha^{-\frac{n-1}{2\sqrt{d_v}}} deg_i(t)\right), & \text{if }n \text{ is odd}
\end{cases}
\end{equation}
where $\phi_{d}$ is a sinusoidal encoding function~\cite{vaswani2017attention} that takes a node degree as an input and returns a degree encoding, and $\alpha$ is a hyperparameter controlling the resolution of degree encoding. 
A larger $\alpha$ smooths out small degree differences, while a smaller $\alpha$ preserves finer details but may introduce noise. 
Here, an index $n$ ranges from 0 to $d_v$-1, where $d_v$ is the node feature dimension.
Note that these structural features change over time both for seen and unseen nodes.
We denote the structural feature augmentation process as $S$, i.e., $\bold{s}_i(t)= S(v_i(t))$.
\begin{simpleexample}
    As shown in Figure~\ref{fig:feature_augmentation}(a), structural features are generated to assign similar features to structurally similar nodes in $\mathcal{G}_{seen}$.
    Note that, at the current time $t^{(9)}$, nodes $v_2$, $v_4$, $v_5$, $v_6$, $v_7$, $v_9$, and $v_{10}$, whose degree is $1$, share the same structural feature vector $\phi_d(1)$.
\end{simpleexample}

\subsubsection{Feature Propagation for Unseen Nodes}

Next, we generate node features for unseen nodes that appear after the training period (i.e., after time $t_{seen}$) while meeting the requirements of CTDGs, i.e., limited access and incremental processing.
For example, in Figure~\ref{fig:feature_augmentation}, $v_{11}$ is an unseen node.

Structural feature augmentation requires only node degrees, which can be incrementally computed, and thus features of unseen nodes can be generated in the same way as for seen nodes, i.e., $\bold{s}_i(t)= \phi_{d}(deg_i(t))$ for any $v_i\notin \mathcal{V}_{seen}$.
This process takes  $O(d_v)$ time for each node independently of the graph size, where $d_v$ is the node feature dimension. 
\begin{simpleexample}
    As shown in Figure~\ref{fig:feature_augmentation}(b), since $v_{11}$ has a degree 0 at time $t^{(9)}$, its structural feature is generated as $\phi_d(0)$. At time $t^{(10)}$, $v_{11}$ participates in $\delta^{(10)}$, increasing its degree to 1 and updating its structural feature to $\phi_d(1)$.
    Similarly, at time $t^{(11)}$, $v_{11}$ participates in $\delta^{(11)}$, raising its degree to 2 and updating its structural feature to $\phi_d(2)$.
\end{simpleexample}

In contrast, random and positional feature augmentations face challenges when applied in the same way to unseen nodes as to seen nodes.
In the case of random feature augmentation, while random features can be assigned to unseen nodes, they may act as noise rather than meaningful absolute positions of the nodes. This is because random features lack any meaningful patterns, and
the trained model (i.e., TGNN) has no chance to directly learn the features.
Moreover, the positional feature augmentation process for seen nodes cannot be directly applied to a CTDG for unseen nodes due to limitations in access and the requirement for incremental processing in CTDGs.

To address these challenges, we propose a method to generate the positional and random features of unseen nodes that align with the feature space of seen nodes.
In essence, our method incrementally propagates the features of seen nodes to unseen nodes through new edges in the input CTDG. 

Specifically, we initialize the node features of each unseen node $v_i \notin  \mathcal{V}_{seen}$ as zero vectors, and in response to each new temporal edge $(v_{i}, v_{j},\bold{x}_{ij}^{(n)},w_{ij}^{(n)}, t^{(n)})$ incident to $v_i$, 
the features of $v_j$ are propagated to $v_i$
to incrementally update its random and positional features as follows:
\vspace{-0.1cm}
\begin{equation}
\label{eq:update_rule1}
\vspace{-0.1cm}
\small
\boldsymbol{r}_i(t^{(n)}) =\frac{deg_i(t^{(n-1)})\boldsymbol{r}_i(t^{(n-1)})+\boldsymbol{r}_j(t^{(n-1)})}{deg_i(t^{(n-1)})+1},
\end{equation}
\begin{equation} \label{eq:update_rule2}
\small
\boldsymbol{p}_i(t^{(n)}) =\frac{deg_i(t^{(n-1)})\boldsymbol{p}_i(t^{(n-1)})+\boldsymbol{p}_j(t^{(n-1)})}{deg_i(t^{(n-1)})+1}.
\end{equation}
Note that this update is applied only to unseen nodes (i.e., only when $v_i \notin  \mathcal{V}_{seen}$).
This update, essentially linear interpolation, has a constant time complexity of $O(d_v)$, where $d_v$ is the feature dimension,
independently of the graph size.
\begin{simpleexample}
    In Figure~\ref{fig:feature_augmentation}(c), we illustrate feature propagation for the unseen node $v_{11}$. Let the random features of seen nodes be $\vecr_1=[0.1, -0.2]$ and $\vecr_2=[0.1, 0.3]$; and let the positional features be $\vecp_1=[0.9, 0.7]$ and  $\vecp_2=[0.7, 0.8]$. Initially, at time $t^{(9)}$, the feature vectors of $v_{11}$ are set to zero, i.e., $\vecr_{11}(t^{(9)})=\vecp_{11}(t^{(9)})=[0,0]$.
    At time $t^{(10)}$, after $v_{11}$ interacts with $v_1$ in $\delta^{(10)}$,  its features are updated to $\vecr_{11}(t^{(10)})=[0.1,-0.2]$ and $\vecp_{11}(t^{(10)})=[0.9,0.7]$ following Eqs.~\eqref{eq:update_rule1} and \eqref{eq:update_rule2}. 
    Similarly, at time $t^{(11)}$, after  $v_{11}$ interacts with $v_2$ in $\delta^{(11)}$, its features are further updated to $\vecr_{11}(t^{(11)})=[0.1,0.05]$ and $\vecp_{11}(t^{(11)})=[0.8,0.75]$ following the same equations.
\end{simpleexample}

\begin{figure*}[t]
    \centering
    \vspace{-2mm}\includegraphics[width=1\linewidth]{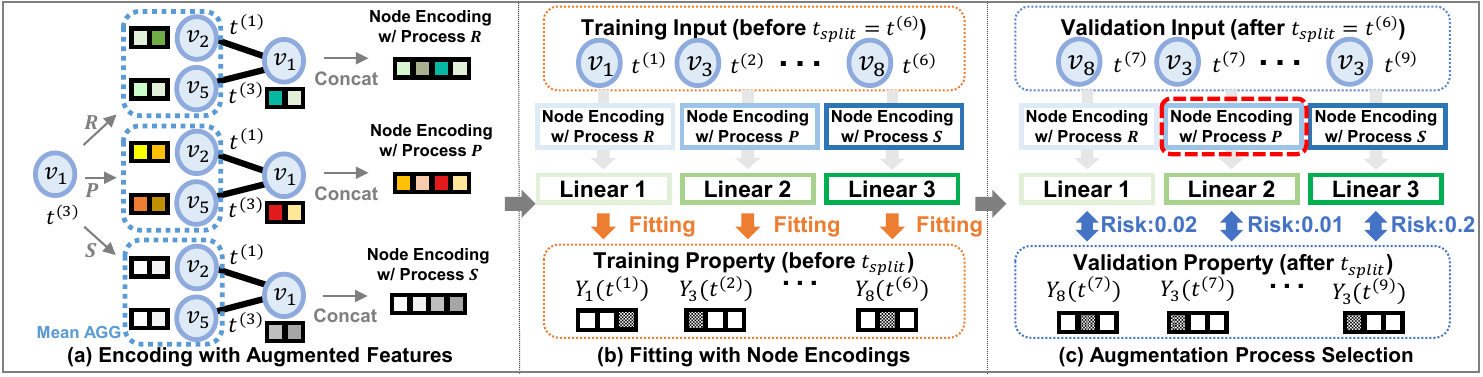}
    \centering
    \vspace{-6mm}
    \caption{ \label{fig:selection_model}
        Overview of node feature selection in \method.
        (a) Based on the information of the target node and its recent neighbors, each node feature augmentation process is applied to generate node encodings.
        (b) 
        For each node feature augmentation process,
        \method performs linear fitting of the corresponding node encodings to the training property set before the split time $t_{split}$.
        (c) \method evaluates the empirical risk for each process based on the validation property set after $t_{split}$ and selects the node augmentation process with minimal risk.
    }
\end{figure*}

\subsection{Node Feature Selection}
\label{sec:method:selection}


The three previously described feature augmentation processes  (i.e., random, structural, and positional) need to be selectively applied, especially under distributional shifts. 
In this subsection, we present our efficient and effective feature selection process for this purpose.

\begin{note}[Relation with Other Components]
Among the candidate augmented node features generated through \textbf{node feature augmentation} (Section~\ref{sec:method:augmentation}), the most effective ones for node property prediction under distribution shifts are selected in this step. These selected ones are then used as input of the \textbf{\model model} (Section~\ref{sec:method:model}) for both training and inference.
\end{note}



In general, most machine learning models are trained to minimize the empirical risk (e.g., cross-entropy loss) on the training set using label supervision, and this training approach is referred to as empirical risk minimization (ERM). 
However, models trained using ERM often exhibit poor performance on the test set under distribution shifts. 
Past studies~\cite{chen2024understanding,degrave2021ai,geirhos2020shortcut} attribute this to the tendency of ERM to learn spurious or shortcut features that are ineffective under distribution shifts.

In CTDGs, distributional shifts can arise due to temporal changes, as discussed in Section~\ref{related:ds}. Thus, it is essential to filter out (augmented) node features that might act as spurious or shortcut features, ensuring TGNNs remain effective under distributional shifts.
Indeed, we demonstrate empirically in Section~\ref{sec:exp:ablation} (refer to Table~\ref{tab:ablation_performance}) that using selective node features achieves better performance than using all features. 

A common approach to feature selection involves training a TGNN model using each node feature individually on a single training set based on ERM.
Then, a standard validation process is conducted to select the node feature, minimizing the empirical risk on the validation set with distribution shifts (i.e., validation loss).
However, this approach is highly inefficient and time-consuming, as it requires repeatedly training and validating TGNNs for each individual node feature.

\subsubsection{Overview}
We propose an efficient feature selection process for CTDGs that does not require repetitive training and validation of TGNN.
Observing that informative features primarily exhibit invariant correlations with labels, which can be identified independently of trained machine learning models, we propose using a linear model instead of TGNNs to accelerate feature selection.
Specifically, a linear model is trained using ERM on the training set, and feature selection is performed based on ERM performance on a validation set with distribution shifts.
Moreover, leveraging the efficiency of the linear model, it becomes feasible to explore multiple training-validation splits with varying degrees of distributional shifts.
The efficiency and effectiveness of our feature selection method are empirically demonstrated in Section~\ref{sec:exp:ablation} (refer to Table~\ref{tab:ablation_performance} and Figure 6 in Online Appendix I~\cite{Lee_SPLASH_Online_Appendix_2024}).

A visual overview of our node feature selection process is provided in Figure~\ref{fig:selection_model}.
The process consists of three stages: (a) encoding nodes with augmented features, (b) training linear models using the encoded features, and (c) selecting node features that minimize empirical risk across various validation sets based on the linear models.
Below, we describe each step.
\subsubsection{Encoding with Augmented Features}
In this stage, information from each node and its neighbors is encoded through various node feature augmentation processes. 
TGNNs~\cite{tgn_icml_grl2020, cong2022we,yu2023towards} typically generate the representation of a node at a specific time using the $k$ most recent temporal edges incident to the node.
Thus, also in \method,
for each node $v_i$ at time $t$, we use the $k$ most recent temporal edges incident to the node, which we denote by  $\mathcal{N}_{i}(t)$.
That is, $\mathcal{N}_{i}(t)$ the most recent $k$ edges when chronologically ordering
\vspace{-0.23cm}
\begin{equation}
\vspace{-0.23cm}
\small
\mathcal{E}_i(t)=\left\{\delta^{(m)} \mid \delta^{(m)} \in \mathcal{G} \wedge v_i \in \delta^{(m)} \wedge t^{(m)}\leq t\right\},
\end{equation}
i.e., the temporal edges incident to $v_i$ up to time $t$.


To encode information from a node and its recent neighbors, we first apply mean aggregation to the features of the recent neighbors and then concatenate the result with the feature of the node itself, avoiding the use of any complex encoder.
That is, 
the encoding of each node $v_i$ at time $t$ for the node features generated by a process $X$ is obtained as follows:
\vspace{-0.1cm}
\begin{equation}
\vspace{-0.1cm}
\small
\boldsymbol{x}_{i}^{E}(t) = \left [\boldsymbol{x}_{i}(t) \Big\Vert \frac{1}{|\mathcal{N}_{i}(t)|} \sum\nolimits_{\delta^{(l)}\in\mathcal{N}_{i}(t)\wedge v_j\in\delta^{(l)}}\boldsymbol{x}_{j}(t^{(l)}) \right],
\end{equation}
where $\Vert$ indicates the concatenation operation, and $v_j$ denotes the neighbor of $v_i$ (i.e., the other endpoint) in temporal edge $\delta^{(l)}$. The features $\boldsymbol{x}_i(t)$ and $\boldsymbol{x}_j(t^{(l)})$ are those generated by the process $X$, i.e., $\boldsymbol{x}_i(t)= X(v_i(t)),$ and $\boldsymbol{x}_j(t^{(l)}) = X(v_j(t^{(l)}))$.

\subsubsection{Fitting with Node Encodings}
In this stage, 
for each feature augmentation process,
a linear model is trained with ERM on the training set, using the corresponding node encodings.
Below, we denote the available set of node properties before the test time $t_{test}$ as follows:
\vspace{-1mm}
\begin{equation}
\vspace{-1mm}
\small
\mathcal{Y}_{A}=\{(v_i, t, Y_i(t)) \mid v_i \in \mathcal{V}, t<t_{test}\},
\end{equation}
where $Y_i(t)$ denotes the label of $v_i$ at time $t$. 
To identify feature augmentation processes that remain effective under distribution shifts,  we simulate a distribution shift scenario by generating a temporal split within $\mathcal{Y}_A$. 
That is, our underlying assumption is that the node features that are effective in this simulated scenario are also effective in the actual test setting, as shown empirically in Section~\ref{sec:exp:experiment_results} (see Table~\ref{tab:ablation_performance}).
To this end, we first divide the available node property set into training and validation property sets chronologically based on the split time $t_{split}$, which is  smaller than $t_{test}$, i.e., 
\begin{equation}
\small
\mathcal{Y}_{T} =\{ (v_i, t, Y_i(t)) \mid v_i \in \mathcal{V}, t\leq t_{split}\}, 
\mathcal{Y}_{V} =\mathcal{Y}_{A}\setminus\mathcal{Y}_{T},
\end{equation}
where $\mathcal{Y}_{T}$ is the training property set up to $t_{split}$, while $\mathcal{Y}_{V}$ is a validation property set after $t_{split}$.
Notably, temporal changes may lead to distribution shifts between the training and validation data, both in node features and labels.

Then, we train a separate linear model for each node feature augmentation process, using the corresponding node encodings as input to predict the node properties.
We use ERM for training, aiming to minimize the empirical risk (spec., the cross-entropy loss) on the training set as follows:
\begin{equation}
\small
\boldsymbol{W}_{X}^{*} = \arg\min_{\mathbf{W}} \frac{1}{\left|\mathcal{Y}_{T} \right|}\sum\nolimits_{(v_i, t, Y_i(t))\in \mathcal{Y}_{T}}\mathcal{L}(\boldsymbol{W}\boldsymbol{x}_{i}^{E}(t), Y_i(t)),
\end{equation}
where $\boldsymbol{W}_{X}^{*}$ denotes the weight of the linear model for node feature augmentation process $X$, trained on $\mathcal{Y}_{T}$.

\subsubsection{Augmentation Process Selection}
After training the linear models for all node feature augmentation processes (i.e., random, positional, and structural processes), the empirical risk on the validation node property set $\mathcal{Y}_{V}$ is measured for each process as follows:
\vspace{-0.23cm}
\begin{equation}
\small
\mathcal{L}_{X}(\mathcal{Y}_{V}|\mathcal{Y}_{T})=\frac{1}{\left|\mathcal{Y}_{V} \right|}\sum\nolimits_{(v_i, t, Y_i(t))\in \mathcal{Y}_{V}}\mathcal{L}(\boldsymbol{W}_{X}^{*}\boldsymbol{x}_{i}^{E}(t), Y_i(t)),
\end{equation}
where $\mathcal{L}_{X}$ denotes the empirical risk of the linear model corresponding to node feature augmentation process $X$ on the validation node property set $\mathcal{Y}_{V}$.
Due to the potential distribution shifts between the training and validation data, the empirical risk on the validation set serves as a useful indicator of performance under such shifts.


To ensure feature selection that is robust across varying degrees of distributional shifts, \method divides the training and validation property sets based on multiple split times,\footnote{In our work, we use five split times, dividing the available property set into training/validation splits of 10/90\%, 30/70\%, 50/50\%, 70/30\%, and 90/10\%.} denoted by $t_{split}^{(1)},t_{split}^{(2)}, ..., t_{split}^{(n')}$, resulting in the set $\mathcal{S}$ of pairs of training and validation property sets, i.e.,
\vspace{-0.1cm}
\begin{equation}
\vspace{-0.1cm}
\small
\mathcal{S}=\{ (\mathcal{Y}_{T}^{(1)},\mathcal{Y}_{V}^{(1)}), (\mathcal{Y}_{T}^{(2)},\mathcal{Y}_{V}^{(2)}), \cdots, (\mathcal{Y}_{T}^{(n')},\mathcal{Y}_{V}^{(n')})\},
\end{equation}
where $\mathcal{Y}^{(n)}_{T}$ and $\mathcal{Y}^{(n)}_{V}$ are training and validation node property sets, split based on split time $t^{(n)}_{split}$.
\method selects the node feature augmentation process that minimizes the sum of empirical risks on the multiple validation property sets in  $\mathcal{S}$:
\begin{equation}
\small
X^{\star} = \arg\min_{X} \sum\nolimits_{(\mathcal{Y}_{T}^{(n)},\mathcal{Y}_{V}^{(n)})\in \mathcal{S}}\mathcal{L}_{X}(\mathcal{Y}_{V}^{(n)}|\mathcal{Y}_{T}^{(n)}),
\end{equation}
where $X^{\star}$ denotes the selected node feature augmentation process.
Note that, here, a linear model is trained separately on each corresponding training property set.
It is also important to note that multiple splits can be considered without much computational cost, since our feature selection is based on simple linear models, rather than TGNNs.


\begin{figure*}[t]
    \centering
    \vspace{-2mm}\includegraphics[width=1\linewidth]{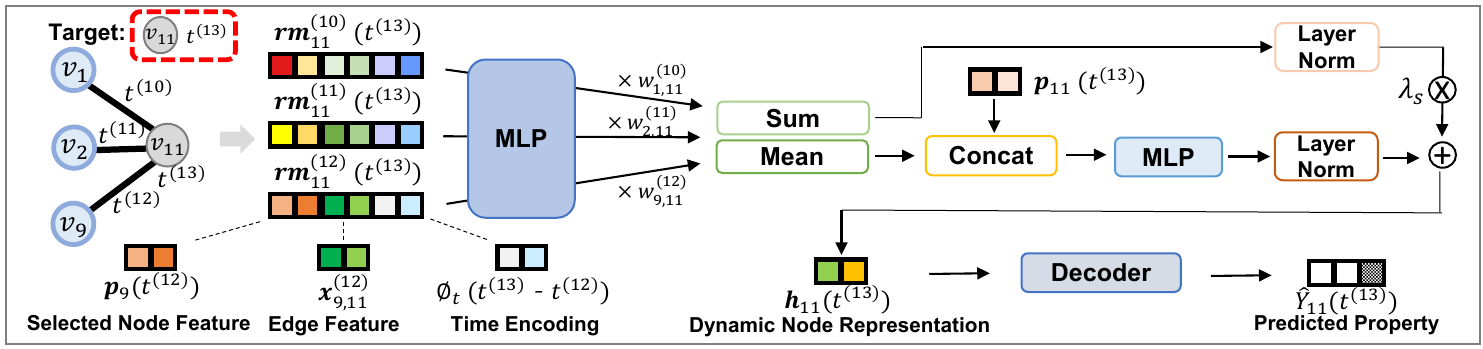}
    \centering
    \vspace{-6mm}
    \caption{ \label{fig:SLIM_model}
        Overview of the \model model, our proposed simple MLP-based TGNN.
        To create the  dynamic node representation of a target node,
        this model utilizes only MLPs with
        the selected augmented node features of the target node and its recent neighbors. The generated node representation is fed into a decoder to predict the property of the target node. 
    }
\end{figure*}

\subsection{SLIM Model}\label{sec:method:model}

 
As discussed in Section~\ref{intro:finding_dist} and empirically confirmed in
Section~\ref{sec:exp:experiment_results} (refer to Figure~\ref{fig:generalization_ratio}), existing TGNNs often exhibit limited generalization capabilities under distribution shifts, primarily due to their complex architectures.
Thus, under distribution shifts, a simpler model can be both more effective and more efficient.
Based on this motivation, in this subsection,
we propose a \model (\underline{S}imple M\underline{L}P-based model with \underline{I}ntegration of \underline{M}essages), a simple TGNN for generating dynamic representations of nodes over time for a given CTDG.

\begin{note}[Relation with Other Components]
The \model model takes the previously selected augmented node features as input, along with the CTDG.
These features are chosen by \textbf{feature selection} (Section~\ref{sec:method:selection}) from candidate augmented node features generated by \textbf{feature augmentation} (Section~\ref{sec:method:augmentation}).
\end{note}



\subsubsection{Overview}

\model is a simple MLP-based model, and it does not rely on complex components, such as self-attention, RNNs, or memory modules, which are commonly used in existing TGNNs.
Despite its simplicity, \model is designed to effectively utilize the proposed augmented node features.

A pictorial description of \model is given in Figure~\ref{fig:SLIM_model}.
Specifically, \model consists of two main modules: (a) the message encoding module and (b) the aggregation module.
In both modules, the selected augmented node feature of every node at time $t$ is generated (or incrementally updated) through the selected node feature augmentation process, i.e., $\boldsymbol{x}^{\star}_m(t)= X^{\star}(v_m(t)), ^{\forall}{v_m}\in \mathcal{V}$ (refer to Section~\ref{sec:method:augmentation} for the feature augmentation processes and Section~\ref{sec:method:selection} for the selection process).
Below, we describe each module.
\begin{simpleexample}
    If the positional feature augmentation process $P$ is selected through the feature selection, as shown in Figures~\ref{fig:selection_model} and~\ref{fig:SLIM_model},  the positional features are used in \model as input, i.e., $\boldsymbol{x}^{\star}_m(t)= P(v_m(t))=\boldsymbol{p}_m(t), ^{\forall}{v_m}\in \mathcal{V}$.
\end{simpleexample}

\subsubsection{Message Encoding Module}
Based on recent temporal edges and incrementally computed augmented node features,
\model computes the latest representation of a given target node, denoted by $v_i$, by aggregating messages from its recent neighbors.
In the message encoding module, we encode the message from each recent neighbor.

Specifically, at time $t$, 
\model generates a raw message from each of the $k$ most recent temporal edges incident to $v_i$
(i.e., from each
$\delta^{(l)}=(v_{i}, v_{j},\bold{x}_{ij}^{(l)},w_{ij}^{(l)}, t^{(l)})$ 
or $(v_{j}, v_{i},\bold{x}_{ji}^{(l)},w_{ji}^{(l)}, t^{(l)})\in\mathcal{N}_{i}(t)$),
as follows:
\vspace{-1mm}
\begin{equation}
\vspace{-1mm}
\small
\vecr\vecm_{i}^{(l)}(t)= [\boldsymbol{x}_{j}^{\star}(t^{(l)})||\boldsymbol{x}_{ij}^{(l)}||\phi_t(t-t^{(l)})],
\end{equation}
where $\vecr\vecm_{i}^{(l)}(t)$ denotes the raw message vector from the recent temporal edge $\delta^{(l)}$ to $v_i$ at time $t$; $\boldsymbol{x}_{j}^{\star}(t^{(l)})$  denotes the selected augmented node feature vector of the neighbor $v_{j}$ at time $t^{(l)}$; and $\boldsymbol{x}_{ij}^{(l)}$ denotes the given edge feature vector of $\delta^{(l)}$.
As $\phi_t$, we employ the following time encoding function~\cite{cong2022we}:
\begin{equation}
\small
\label{temporal_enc}
    \phi_t(t')  = cos\left(t' \cdot [\alpha^{-\frac{0}{\beta}} \vert\vert  \alpha^{-\frac{1}{\beta}}\vert\vert \cdots \vert\vert \alpha^{-\frac{d_{t}-1}{\beta}}]\right), 
\end{equation}
where $d_{t}$ denotes the dimension of time encoding vectors; and
scalars $\alpha$, $\beta$ and $d_{t}$ are hyperparameters.

Each raw message is then transformed into a message by an MLP for message encoding, denoted as $MLP_1$, as follows:
\begin{equation}
\small\vecm_{i}^{(l)}(t)= MLP_1(\vecr\vecm_{i}^{(l)}(t)) \times w_{ij}^{(l)},
\end{equation}
where $w_{ij}^{(l)}$ denotes the edge weight in the temporal edge $\delta^{(l)}$.\footnote{If the dataset does not contain edge features, edge feature is excluded from a message, and if there are no explicit edge weights, a weight of one is used.}
Note that $\vecm_{i}^{(l)}(t)$ denotes the message vector from (the other endpoint of) the recent temporal edge $\delta^{(l)}$ to $v_i$ at time $t$.

The process of encoding messages from the $k$ most recent temporal edges takes $O(k((d_v+d_e+d_t)d_h+L_{E}d_h^{2}))$ time independently of the graph size, where $d_v$, $d_e$, and $d_t$ denote the dimensions of node features, edge features, and time encodings respectively; $d_h$ denotes both the hidden dimension of $MLP_1$ and message dimension; and $L_E$ denotes the number of layers in $MLP_1$.

\subsubsection{Aggregation Module}
In this module, 
\model computes the latest representation of a given target node $v_i$ by aggregating the messages encoded in the previous module and combining it with the feature of the target node itself.

First, the intermediate representation $\widetilde{\vech}_{i}$ of the target node 
$v_i$ at time $t$ is obtained by (1) mean aggregating all messages from recent neighbors, (2) combining the result with the feature of the target node itself, and (3) applying an MLP, denoted as $MLP_2$, as follows:
\begin{equation}
\small
\widetilde{\vech}_{i}(t)= MLP_2([\boldsymbol{x}_i^{\star}(t)||\frac{1}{\left|\mathcal{N}_{i}(t) \right|}\sum\nolimits_{\delta^{(l)}\in \mathcal{N}_{i}(t)}\vecm_{i}^{(l)}(t) ]),
\end{equation}

Lastly, layer normalization~\cite{ba2016layer}, which is known to enhance generalization~\cite{xu2019understanding}, is applied to the intermediate representation, incorporating a skip connection~\cite{simonyan2014very} through sum aggregation of the messages, to produce the latest representation $\vech_{i}(t)$ of the target node $v_i$ as follows:
\begin{equation}
\small
\vech_{i}(t)= LN_{1}(\widetilde{\vech}_{i}(t))+\lambda_{s} LN_{2}(\sum\nolimits_{\delta^{(l)}\in \mathcal{N}_{i}(t)}\vecm_{i}^{(l)}(t)),
\end{equation}
where 
$LN_1$ and $LN_2$ are layer normalization functions; and $\lambda_s$ is the skip connection weight, which is a hyperparameter.

The process in the aggregation module takes $O((k+d_v)d_h+L_{A}d_h^2)$ time for each node regardless of the graph size, where $d_v$ denotes the node feature dimension; $d_h$ denotes both the dimension of node representations and the hidden dimension of $MLP_2$; and $L_A$ denotes the number of layers in $MLP_2$.

\subsubsection{Prediction and Training Processes}
Similar to other TGNNs, \method feeds the representation of a node at time $t$
into a decoder to predict its property at time $t$ as follows:
\vspace{-1mm}
\begin{equation}
\vspace{-1mm}
\small
 \widehat{Y}_i(t)=Decoder(\vech_{i}(t)),
\end{equation}
where $\widehat{Y}_i(t)$ is the predicted node property of $v_i$ at time $t$, and as $Decoder$ we employ an MLP, a common choice in TGNNs.

The prediction process takes $O(d_hd_l+L_{d}d_h^2)$ time, where $d_h$ denotes both the dimension of node representations and the hidden dimension of $Decoder$; $d_l$ denotes the dimension of the predicted node property; and $L_{d}$ denotes the number of layers in $Decoder$.
Note that all processes (spec., message encoding, message aggregation, and prediction process) involved in predicting the node property of each node using trained \model take time independent of the overall graph size.

In the training phase, MLPs in the message encoding module, the aggregation module, and the decoder are trained to minimize the empirical risk on the training set (i.e., temporal edges in the training period) as follows:
\vspace{-1mm}
\begin{equation}
\vspace{-1mm}
\small
\mathcal{L}_{train}=\frac{1}{\left|\mathcal{Y}_{train} \right|}\sum\nolimits_{(v_i, t, Y_i(t))\in \mathcal{Y}_{train}}\mathcal{L}(\widehat{Y}_i(t), Y_i(t)),
\end{equation}
where $\mathcal{Y}_{train}$ denotes the training node property set before $t_{seen}$.
Once trained, \model and the decoder can be used to incrementally predict the dynamic property of each node in the input CTDG.

\begin{table}[!t]
    \centering
    \vspace{-1.5mm}
    \caption{Statistics of datasets used in our experiments.
    Since node property queries are independent of edge appearances, the number of edges and property queries can differ.}
    \scalebox{0.82}{
        \begin{tabular}{lccccccc}
            \toprule
             & \multicolumn{3}{c}{\textbf{\shortstack{Dynamic Anomaly \\ Detection}}} & \multicolumn{2}{c}{\textbf{\shortstack{Dynamic Node \\ Classification}}} & \multicolumn{2}{c}{\textbf{\shortstack{Node Affinity \\ Prediction}}} \\ 
            \cmidrule(lr){2-4} \cmidrule(lr){5-6} \cmidrule(lr){7-8}
             & \textbf{Reddit} & \textbf{Wiki} & \textbf{MOOC} & \makecell[c]{\textbf{\shortstack{Email\\-EU}}} & \textbf{GDELT} & \makecell[c]{\textbf{\shortstack{TGBN\\-trade}}} & \makecell[c]{\textbf{\shortstack{TGBN\\-genre}}} \\
            \midrule
            \# nodes & 10,984 & 9,227 & 7,047 & 986 & 6,829 & 255 & 1,505 \\ 
            \# edges & 672k & 157k & 412k & 332k & 1,913k & 468k & 17,858k \\ 
            \# queries & 672k & 157k & 412k & 201k & 438k & 7k & 256k \\
            node feats & no & no & no & no & yes & no & no \\ 
            edge feats & yes & yes & yes & no & yes & no & no \\ 
            edge weight & no & no & no & no & no & yes & yes \\
            $d_v$ & N/A & N/A & N/A & N/A & 413 & N/A & N/A \\
            $d_e$ & 172 & 172 & 4 & N/A & 182 & N/A & N/A \\
            \# label & 2 & 2 & 2 & 42 & 81 & 255 & 513 \\ 
            \bottomrule
        \end{tabular}
    }
    \label{tab:dataset}
\end{table}

\section{Experiments}
\label{sec:exp}

In this section, we review our experiments regarding accuracy \& generalization, efficiency \& scalability, ablation study, robustness to distribution shifts, and qualitative analysis.

\subsection{Experiment Details}
\label{sec:exp:details}
In this subsection, we outline datasets for each subtask in node property prediction, baseline methods, and evaluation metrics used in our experiments. 

\smallsection{Datasets for Dynamic Anomaly Detection}
We assess the performance of \method on three real-world datasets (Wikipedia, Reddit, and MOOC~\cite{kumar2019predicting}) for dynamic anomaly detection, where the node property is each user's state, indicating whether it is normal or abnormal at a given time.

\smallsection{Datasets for Dynamic Node Classification}
We evaluate the performance of \method on two real-world datasets (Email-EU~\cite{paranjape2017motifs} and GDELT~\cite{zhou2022tgl}) for dynamic node classification.
In them, the node property is each user's class at a given time.

\smallsection{Datasets for Node Affinity Prediction}
We evaluate the performance of \method on two real-world datasets (TGBN-trade, TGBN-genre~\cite{huang2024temporal}) for node property prediction.
In these datasets, the node property is each node's future affinity of the next time step to the subset of other nodes.

Note that these property labels from all datasets are inherently given in the original datasets. 
Some dataset statistics are provided in Table~\ref{tab:dataset}, and 
across all datasets, 
we utilize the chronological 10/10/80$\%$ split for training, validation, and test sets. 
Refer to Online Appendix A~\cite{Lee_SPLASH_Online_Appendix_2024} for details of these real-world datasets.

\smallsection{Synthetic Datasets with Artificial Distribution Shifts}
We create three synthetic datasets, Synthetic-50/70/90, with shift intensities of 50, 70, and 90, respectively, to evaluate robustness under varying distribution shifts. Higher shift intensity corresponds to a greater degree of shift. A detailed description is provided in Online Appendix B~\cite{Lee_SPLASH_Online_Appendix_2024}.

\smallsection{Baselines Methods and Evaluation Metric}
We extensively compare \method with several TGNN methods capable of predicting node property on edge streams under distribution shifts.
In the case of existing TGNN models (JODIE~\cite{kumar2019predicting}, DySAT~\cite{sankar2020dysat}, TGAT~\cite{xu2020inductive}, TGN~\cite{tgn_icml_grl2020}, GraphMixer~\cite{cong2022we}, DyGFormer~\cite{yu2023towards}, FreeDyG~\cite{tian2023freedyg}, and SLADE\footnote{SLADE is specifically designed for dynamic anomaly detection and is evaluated exclusively on this task.}~\cite{lee2024slade}), no specific node features are provided as input when node features are absent.
In addition, we include additional baselines, denoted as baseline+$RF$, that employ random features, which are straightforward augmented features, as node features for all nodes, including unseen nodes, in existing TGNN models.
For the robustness to distribution shift experiment, we include DTDG-based methods~\footnote{
Note that these DTDG-based models face challenges when applied to real-world datasets, as they are limited to predicting a single static property label per node for each graph snapshot, and they can't provide real-time solutions.} for handling distribution shifts, such as DIDA~\cite{zhang2022dynamic} and SLID~\cite{zhang2024spectral}.
For every baseline, we train each model and a decoder using train sets and do hyperparameter tuning with validation sets.
A detailed description of the implementation is provided in Online Appendix F~\cite{Lee_SPLASH_Online_Appendix_2024}.

To evaluate the performance of each model, we employ the Area Under ROC (AUC) for dynamic anomaly detection, the F1 Score for dynamic node classification, and NDCG@10 for node affinity prediction.
Higher values of these metrics indicate better performance.
A detailed description of evaluation metrics is provided in Online Appendix E~\cite{Lee_SPLASH_Online_Appendix_2024}.

\begin{table}[!t]
    \vspace{-2mm}
    \centering
    \caption{\label{tab:main_performance} Performance (in \%) in the prediction of node properties. 
    For each dataset, the best and the second-best performances are highlighted in \textbf{boldface} and \ul{underlined}, respectively.
    In most cases, \method performs best compared to other baselines.
    }
    \setlength{\tabcolsep}{2.5pt}
    \small
    \scalebox{0.75}{
        \def\arraystretch{1.0}
        \begin{tabular}{@{}l|ccc|cc|cc@{}}
        \toprule
         & \multicolumn{3}{c}{\textbf{\shortstack{Dynamic Anomaly \\ Detection}}} & \multicolumn{2}{c}{\textbf{\shortstack{Dynamic Node \\ Classification}}} & \multicolumn{2}{c}{\textbf{\shortstack{Node Affinity \\ Prediction}}} \\ 
        & \multicolumn{3}{c}{\textbf{(AUC)}} & \multicolumn{2}{c}{\textbf{(F1 Score)}} & \multicolumn{2}{c}{\textbf{(NDCG@10)}} \\
         \cmidrule(lr){2-4} \cmidrule(lr){5-6} \cmidrule(lr){7-8}
         & \textbf{Reddit} & \textbf{Wiki} & \textbf{MOOC} & \makecell[c]{\textbf{\shortstack{Email\\-EU}}} & \textbf{GDELT} & \makecell[c]{\textbf{\shortstack{TGBN\\-trade}}} & \makecell[c]{\textbf{\shortstack{TGBN\\-genre}}} \\ \midrule
        JODIE~\cite{kumar2019predicting} & 55.2 \std{(1.0)} & 80.6 \std{(0.6)} & 62.8 \std{(1.0)} & 10.5 \std{(0.3)} & 21.1 \std{(0.2)} & 35.2 \std{(0.4)} & 35.6 \std{(0.0)} \\
        DySAT~\cite{sankar2020dysat} & 56.9 \std{(2.3)} & 80.6 \std{(0.5)} & 64.2 \std{(0.6)} & 11.6 \std{(0.0)} & 21.5 \std{(0.1)} & 35.1 \std{(0.4)} & 35.7 \std{(0.1)} \\
        TGAT~\cite{xu2020inductive} & 61.0 \std{(0.9)} & 79.1 \std{(1.3)} & 62.0 \std{(2.6)} & 9.8 \std{(1.3)} & 12.8 \std{(0.7)} & 35.1 \std{(0.3)} & 35.6 \std{(0.0)} \\
        TGN~\cite{tgn_icml_grl2020} & 59.3 \std{(0.4)} & 80.4 \std{(1.5)} & 70.1 \std{(0.6)} & 11.2 \std{(2.7)} & 11.6 \std{(0.3)} & 34.7 \std{(0.0)} & 38.6 \std{(0.3)} \\
        GraphMixer~\cite{cong2022we} & 63.0 \std{(1.5)} & 83.8 \std{(0.8)} & 66.1 \std{(1.4)} & 11.9 \std{(2.2)} & 18.4 \std{(0.4)} & 26.5 \std{(1.3)} & 35.9 \std{(0.0)} \\
        DyGFormer~\cite{yu2023towards} & 63.7 \std{(0.7)} & 84.3 \std{(0.6)} & \ul{71.2} \std{(0.7)} & 14.7 \std{(2.5)} & 20.3 \std{(0.5)} & 34.2 \std{(0.2)} & 37.4 \std{(0.0)} \\
        FreeDyG~\cite{tian2023freedyg} & 65.5 \std{(3.4)} & \textbf{85.8} \std{(0.2)} & 68.4 \std{(1.6)} & 14.7 \std{(1.7)} & 9.7 \std{(0.0)} & 22.6 \std{(1.7)} & 35.3 \std{(0.6)} \\
        SLADE~\cite{lee2024slade} & 52.1 \std{(0.4)} & 84.8 \std{(0.1)} & 62.9 \std{(0.5)} & N/A & N/A & N/A & N/A \\
        \midrule
        JODIE+$RF$ & 48.7 \std{(0.6)} & 79.8 \std{(1.6)} & 60.4 \std{(0.9)} & 93.1 \std{(0.2)} & \ul{24.4} \std{(0.1)} & 44.0 \std{(0.7)} & 37.3 \std{(0.2)} \\
        DySAT+$RF$ & 53.5 \std{(1.5)} & 71.3 \std{(0.9)} & 58.3 \std{(0.9)} & 93.2 \std{(0.1)} & 23.6 \std{(0.3)} & 48.6 \std{(0.0)} & 38.1 \std{(0.1)} \\
        TGAT+$RF$ & 56.1 \std{(4.1)} & 72.6 \std{(2.0)} & 62.2 \std{(0.7)} & \ul{93.3} \std{(0.3)} & 23.0 \std{(0.4)} & \ul{48.7} \std{(0.1)} & 41.2 \std{(0.1)} \\
        TGN+$RF$ & 53.9 \std{(2.1)} & 78.8 \std{(2.3)} & 66.0 \std{(1.0)} & 92.6 \std{(0.5)} & 22.3 \std{(0.4)} & 46.8 \std{(0.6)} & 42.5 \std{(0.3)} \\
        GraphMixer+$RF$ & 51.4 \std{(2.5)} & 69.9 \std{(2.5)} & 59.0 \std{(3.3)} & 83.6 \std{(2.8)} & 18.2 \std{(0.3)} & 26.7 \std{(0.4)} & 36.7 \std{(0.1)} \\
        DyGFormer+$RF$ & 64.1 \std{(0.3)} & 84.5 \std{(0.8)} & 68.5 \std{(3.4)} & 65.7 \std{(3.5)} & 20.8 \std{(0.4)} & 35.2 \std{(1.4)} & \ul{44.1} \std{(0.4)} \\
        FreeDyG+$RF$ & \ul{66.9} \std{(0.4)} & 84.8 \std{(0.7)} & 68.0 \std{(2.3)} & 60.1 \std{(1.3)} & 14.0 \std{(0.4)} & 34.2 \std{(0.1)} & 41.7 \std{(0.4)} \\
        SLADE+$RF$ & 59.0 \std{(0.5)} & 83.6 \std{(0.4)} & 63.2 \std{(0.7)} & N/A & N/A & N/A & N/A \\
        \midrule
        \textbf{\method} & \textbf{73.6} \std{(0.3)} & \ul{84.9} \std{(0.7)} & \textbf{71.5} \std{(0.4)} & \textbf{98.4} \std{(0.1)} & \textbf{25.2} \std{(0.1)} & \textbf{55.3} \std{(0.8)} & \textbf{44.4} \std{(0.2)} \\
        \bottomrule
        \end{tabular}
    }
\end{table}

\subsection{Experimental Results}
\label{sec:exp:experiment_results}

\smallsection{Accuracy \& Generalization}
As shown in Table~\ref{tab:main_performance}, \method significantly outperforms other baseline methods in almost every dataset.
There are two notable observations in this analysis related to the findings in Section~\ref{sec:related:findings}.

First, existing TGNNs (i.e., JODIE, DySAT, TGAT, TGN, GraphMixer, DyGFormer, FreeDyG) without node features are generally ineffective in node property prediction except for dynamic anomaly detection, while simply utilizing random features generally leads to significant performance enhancement.
This result demonstrates that, across various subtasks in node property prediction, node properties are closely related to the additional (positional or structural) information of nodes in CTDGs.
For results of the baselines with selected augmented node features, refer to Online Appendix G~\cite{Lee_SPLASH_Online_Appendix_2024}.

Second, it is noteworthy that under distribution shifts, a simple MLP-based model with selected augmented node features outperforms other baselines, achieving performance gains of up to 13.55$\%$ (in the TGBN-trade dataset) compared to the second-best performing baseline.
This result implies that the model in \method can demonstrate better generalization capabilities than other TGNNs under the distribution shifts.

\begin{figure}[!t]
    \vspace{-2mm}
    \centering
    \includegraphics[width=0.48\textwidth]
    {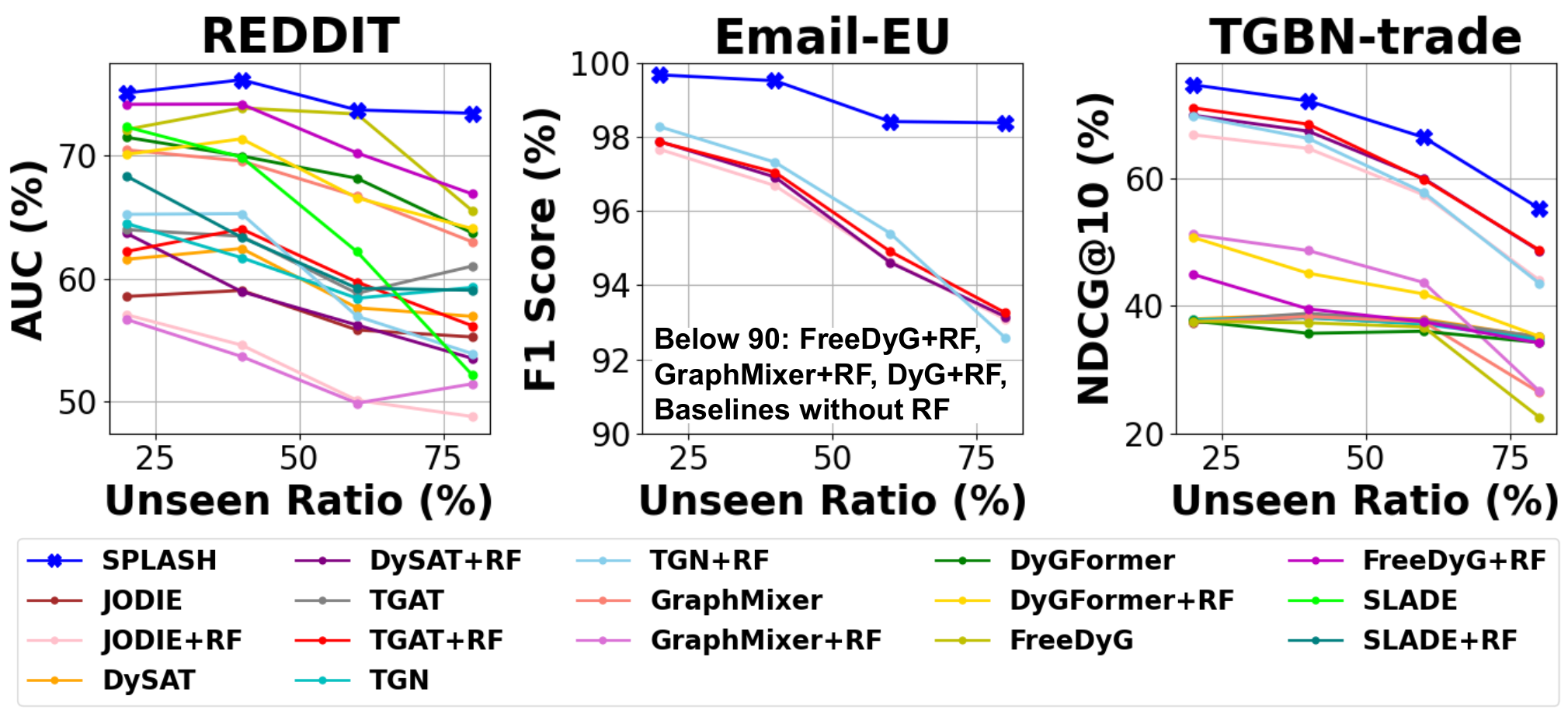}
    \vspace{-3mm}
    \caption{ \label{fig:generalization_ratio}
        Performance (in \%) when varying the ratio of properties unseen during training.
      The latest 10\% of the seen properties, in chronological order, are used for validation, while the earlier properties are used for training.
        Note that \method performs best regardless of the unseen ratio.
    }
\end{figure}

In addition, we measure the performances of the methods while varying the proportion of the unseen part.
Specifically, we utilize the first $90-\calT\%$ of the properties as a train set, the next $10\%$ of the properties as a validation set, and assess each model by using the remaining $\calT\%$ of properties.
We refer $\calT$ as an unseen ratio, where a larger $\calT$ indicates a stronger distribution shift.
As shown in Figure~\ref{fig:generalization_ratio}, \method consistently outperforms all baseline methods across all unseen ratios. Additionally, in most cases, as the distribution shift intensifies (when the unseen ratio increases), the performance gap between the second-best performing baselines and \method increases up to 3.66$\times$ (in the Email-EU dataset). 


\begin{figure}[!t]
    \vspace{-4mm}
    \centering
    \setlength\aboverulesep{0.5pt}
    \setlength\belowrulesep{0.5pt}
    \centering
    \includegraphics[width=1\linewidth]{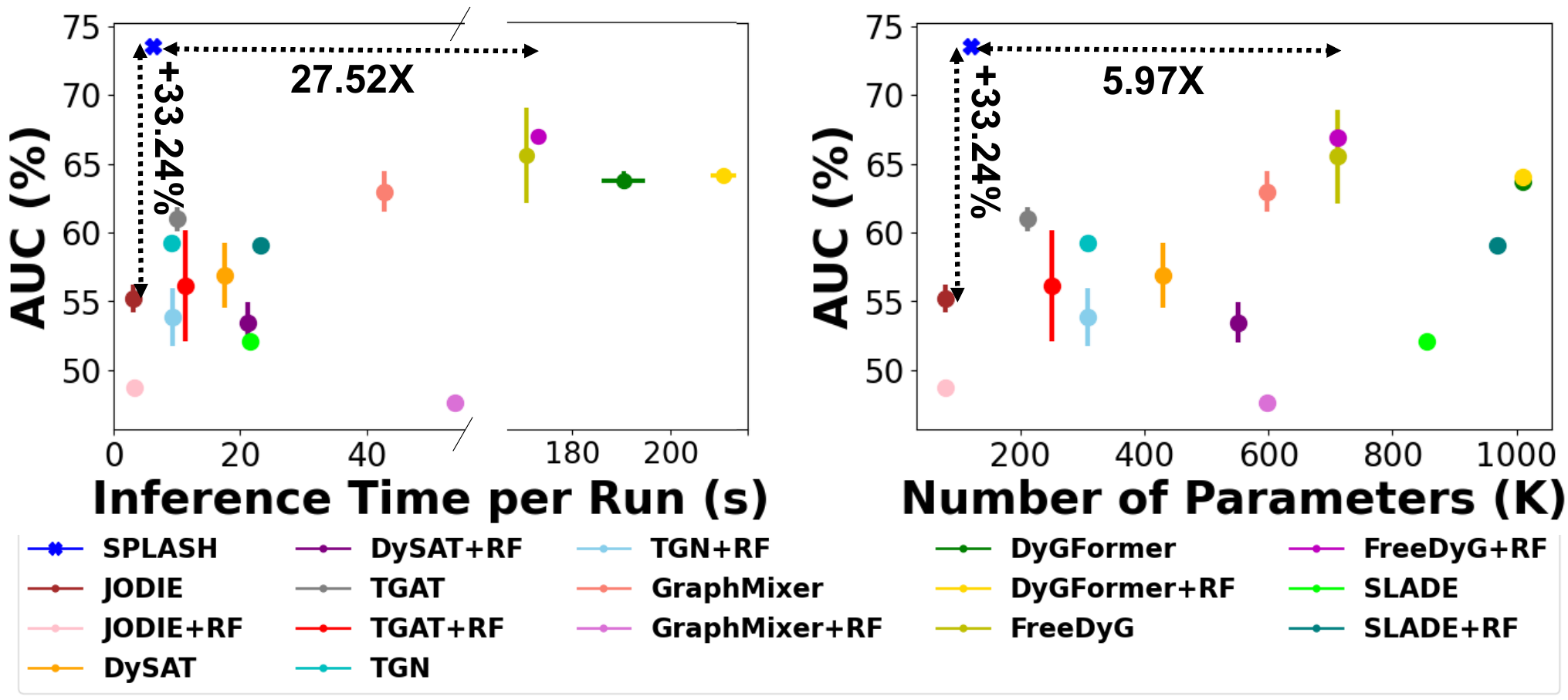} 
    \vspace{-5mm}
    \caption{
    \label{fig:efficiency}
         The left figure shows trade-offs between inference time and AUC, and the right figure shows trade-offs between model size and AUC in the Reddit dataset.
         \method provides the best trade-off not only between speed and performance but also between model size and performance.
    }
\end{figure}

\begin{figure}[!t]
    \vspace{-4mm}
    \centering
    \setlength\aboverulesep{0.5pt}
    \setlength\belowrulesep{0.5pt}
    \centering
    \includegraphics[width=1\linewidth]{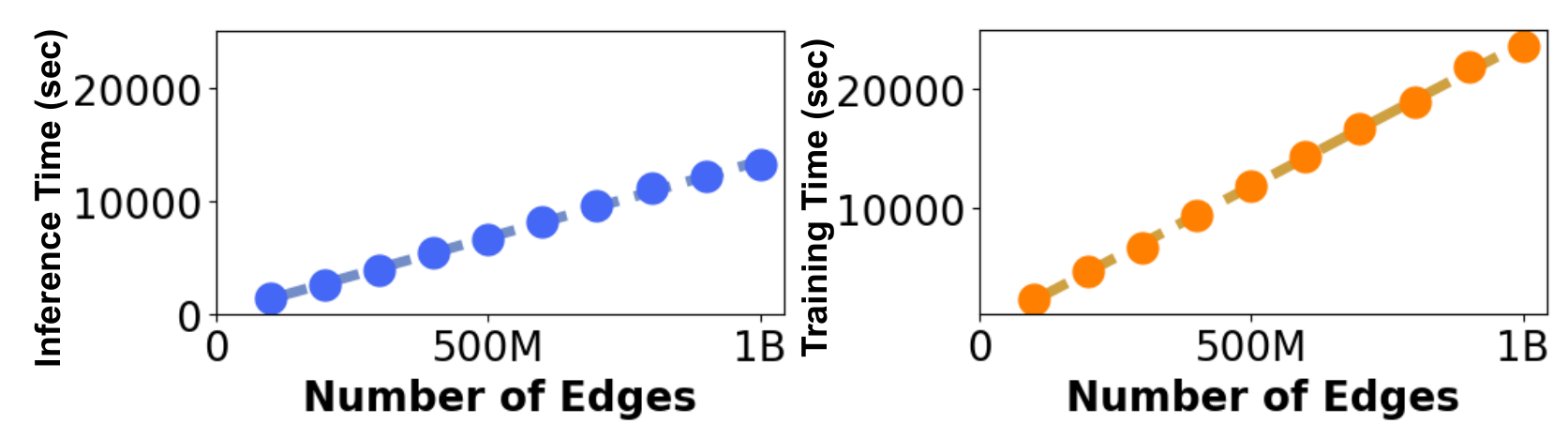} 
    \vspace{-7mm}
    \caption{
    \label{fig:scale_exp}
         Inference time and training time \method relative to the number of edges in the input CTDG. \method demonstrates nearly linear scalability in both inference and training.
    }
\end{figure}

\smallsection{Efficiency \& Scalability}
\label{sec:exp:efficiency}
To evaluate how efficiently \method addresses node property prediction, we measure the trade-off between performance and inference time, as well as performance and the number of parameters, compared with other baselines\footnote{Note that for DyGFormer, GraphMixer, and FreeDyG, their code was adapted from different libraries, which may impact inference time.} in the Reddit dataset.
According to Figure~\ref{fig:efficiency}, \method is 27.52$\times$ faster and 5.97$\times$ lighter than the second-best performing baseline, FreeDyG+$RF$.
\method also significantly outperforms JODIE, the fastest and lightest model, with a 33.24$\%$ performance gain.
For results on training time, refer to Online Appendix H~\cite{Lee_SPLASH_Online_Appendix_2024}.


We evaluate the scalability of \method by measuring its inference and training time synthetic datasets with 100M to 1B edges and 10K to 100K nodes.
Each edge is associated with a label query so that the total number of queries matches the number of edges. 
As shown in Figure~\ref{fig:scale_exp}, both inference and training times scale nearly linearly with the number of edges.
That is, \method processes each edge and query with a time complexity that is independent of the graph size. 

\begin{table}[!t]
    \vspace{-2mm}
    \centering
    \caption{\label{tab:ablation_performance} Performance (in \%) in the prediction of node properties of \method and its variants. 
    For each dataset, the best performances are highlighted in \textbf{boldface}.
    In every case, \method outperforms other variants.
    }
    \setlength{\tabcolsep}{2.5pt}
    \small
    \scalebox{0.75}{
        \def\arraystretch{1.0}
        \begin{tabular}{@{}l|ccc|cc|cc@{}}
        \toprule
         & \multicolumn{3}{c}{\textbf{\shortstack{Dynamic Anomaly \\ Detection}}} & \multicolumn{2}{c}{\textbf{\shortstack{Dynamic Node \\ Classification}}} & \multicolumn{2}{c}{\textbf{\shortstack{Node Affinity \\ Prediction}}} \\ 
        & \multicolumn{3}{c}{\textbf{(AUC)}} & \multicolumn{2}{c}{\textbf{(F1 Score)}} & \multicolumn{2}{c}{\textbf{(NDCG@10)}} \\
         \cmidrule(lr){2-4} \cmidrule(lr){5-6} \cmidrule(lr){7-8}
         & \textbf{Reddit} & \textbf{Wiki} & \textbf{MOOC} & \makecell[c]{\textbf{\shortstack{Email\\-EU}}} & \textbf{GDELT} & \makecell[c]{\textbf{\shortstack{TGBN\\-trade}}} & \makecell[c]{\textbf{\shortstack{TGBN\\-genre}}} \\ \midrule
        SLIM+$ZF$ &  63.2 \std{(0.4)} & 79.3 \std{(0.1)} & 60.3 \std{(0.1)} & 10.9 \std{(0.3)} & 11.7 \std{(0.1)} & 35.0 \std{(0.3)} & 36.1 \std{(0.0)} \\
        SLIM+$RF$ & 61.3 \std{(0.8)} & 78.7 \std{(2.0)} & 66.3 \std{(0.4)} & 95.3 \std{(1.3)} & 24.2 \std{(0.2)} & 55.1 \std{(0.1)} & 42.6 \std{(0.3)} \\
        \midrule
        SLIM+Process $R$ & 62.3 \std{(1.3)} & 79.7 \std{(1.4)} & 66.2 \std{(0.4)} & 98.1 \std{(0.1)} & 24.1 \std{(0.2)} & \textbf{55.3} \std{(0.8)} & 43.7 \std{(0.3)} \\
        SLIM+Process $P$ & 61.1 \std{(1.5)} & 82.2 \std{(0.6)} & 61.9 \std{(0.8)} & \textbf{98.4} \std{(0.1)} & \textbf{25.2} \std{(0.1)} & 51.9 \std{(0.2)} & \textbf{44.4} \std{(0.2)} \\
        SLIM+Process $S$ & \textbf{73.6} \std{(0.3)} & \textbf{84.9} \std{(0.7)} & \textbf{71.5} \std{(0.4)} & 9.3 \std{(0.4)} & 10.9 \std{(0.2)} & 35.1 \std{(0.4)} & 35.4 \std{(0.2)} \\
        \midrule
        SLIM+Joint & 67.2 \std{(2.0)} & 83.0 \std{(1.1)} & 66.1 \std{(0.4)} & 98.1 \std{(0.2)} & 24.0 \std{(0.1)} & 54.5 \std{(0.3)} & 44.0 \std{(0.2)} \\
        \midrule
        \textbf{\method} & \textbf{73.6} \std{(0.3)} & \textbf{84.9} \std{(0.7)} & \textbf{71.5} \std{(0.4)} & \textbf{98.4} \std{(0.1)} & \textbf{25.2} \std{(0.1)} & \textbf{55.3} \std{(0.8)} & \textbf{44.4} \std{(0.2)} \\
        \bottomrule
        \end{tabular}
    }
\end{table}

\smallsection{Ablation Study}
\label{sec:exp:ablation}
%
For the ablation study,
we evaluate cases where (1) zero and random features are used as node features (SLIM+$ZF$, SLIM+$RF$), (2) each of the proposed augmented node features is used without feature selection (SLIM+Process $R,P,S$), and (3) all proposed augmented features are used jointly (SLIM+Joint).
As evident from Table~\ref{tab:ablation_performance}, \method outperforms SLIM+$ZF$ and SLIM+$RF$, which do not utilize the proposed augmented node features and the proposed feature selection process across all datasets.
Notably, compared to SLIM+$ZF$, \method demonstrates an average performance gain of 149.03$\%$ across all datasets.
Moreover, utilizing selected augmented node features demonstrates better performance across all datasets than SLIM+Joint, which uses all augmented node features jointly.
Finally, \method effectively selects optimal augmented node features based on simple linear models (see Online Appendix I~\cite{Lee_SPLASH_Online_Appendix_2024} for their efficiency).

\begin{figure}[!t]
    \vspace{-3mm}
    \centering
    \setlength\aboverulesep{0.5pt}
    \setlength\belowrulesep{0.5pt}
    \centering
    \includegraphics[width=1\linewidth]{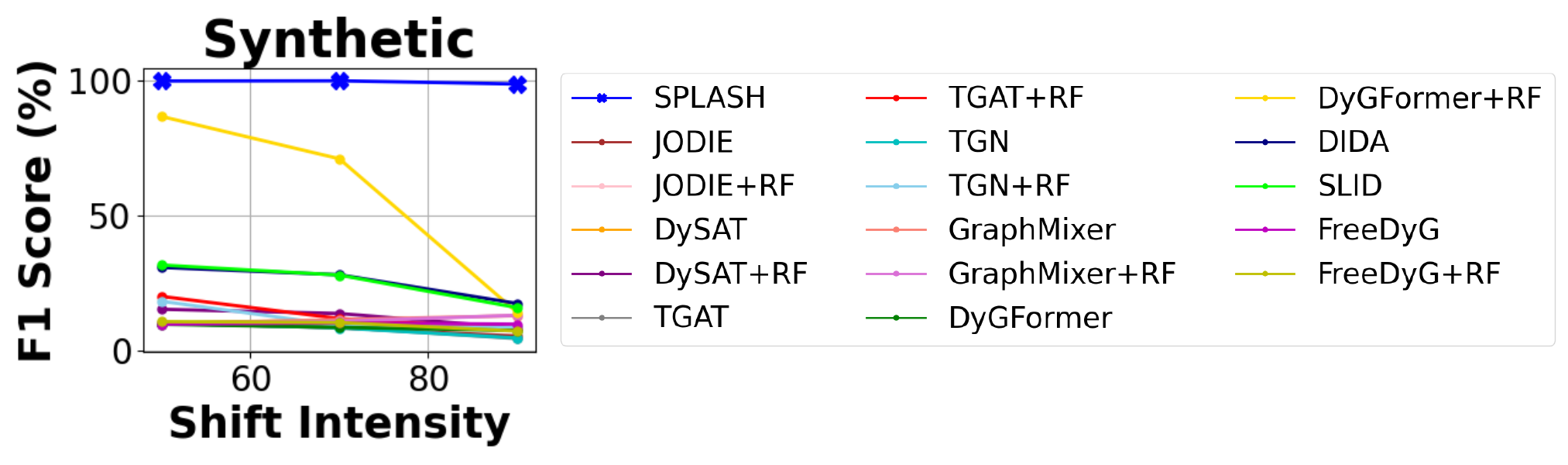} 
    \vspace{-7mm}
    \caption{
    \label{fig:robust_exp}
         Performance (in \%) under varying distribution-shift intensities. \method performs best regardless of the intensity, showing its robustness.
    }
\end{figure}

\smallsection{Robustness to Distribution Shifts}
\label{synthetic_data_exp}
We evaluate the node property prediction performance of SPLASH and various baselines on the synthetic datasets under the artificial distribution shift.
As shown in Figure~\ref{fig:robust_exp}, \method exhibits strong robustness across varying distribution-shift intensities, achieving up to 466.23\% performance gains (in Synthetic-90) over the second-best baseline.
Interestingly, most TGNN models struggle at a shift intensity of 50, showing that even simple positional distribution shifts, such as the appearance of unseen nodes, can significantly degrade property prediction without informative node features. 
While DyGFormer+$RF$ performs similarly to \method at low shift intensity, their gap widens as the intensity grows. 

\begin{figure}[!t]
    \vspace{-2mm}
    \centering
    \setlength\aboverulesep{0.5pt}
    \setlength\belowrulesep{0.5pt}
    \centering
    \includegraphics[width=1\linewidth]{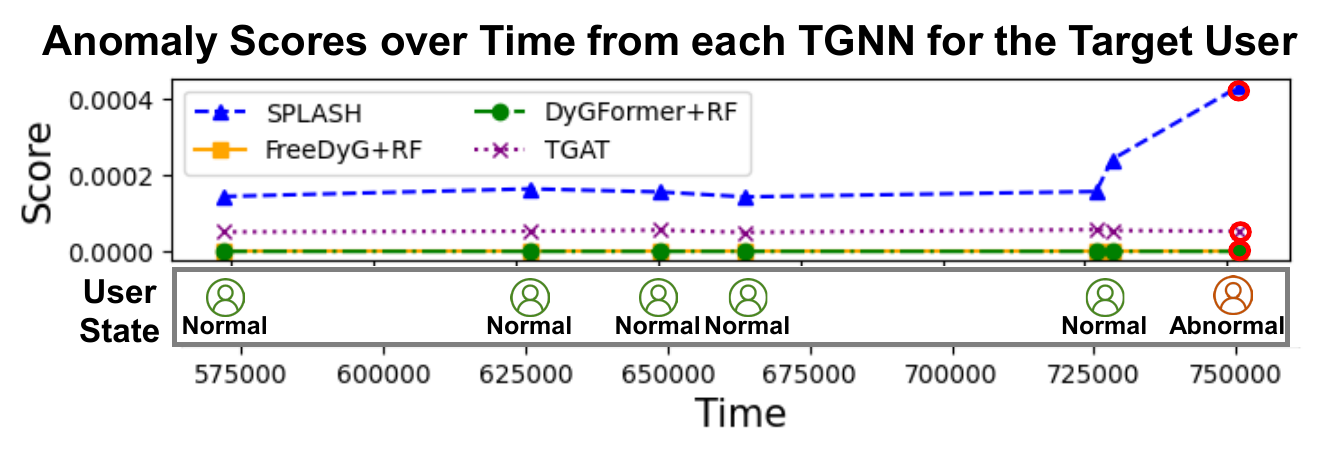} 
    \vspace{-7mm}
    \caption{
    \label{fig:case_study}
         The anomaly scores over time (top) predicted by \method and three baselines and ground-truth dynamic states (bottom) for a specific user in the Reddit dataset.
         \method accurately detects moments of change in the user's dynamic states, which are overlooked by the baseline methods.
    }
\end{figure}

\begin{figure}[!t]
    \vspace{-3mm}
    \centering
    \setlength\aboverulesep{0.5pt}
    \setlength\belowrulesep{0.5pt}
    \centering
    \includegraphics[width=1\linewidth]{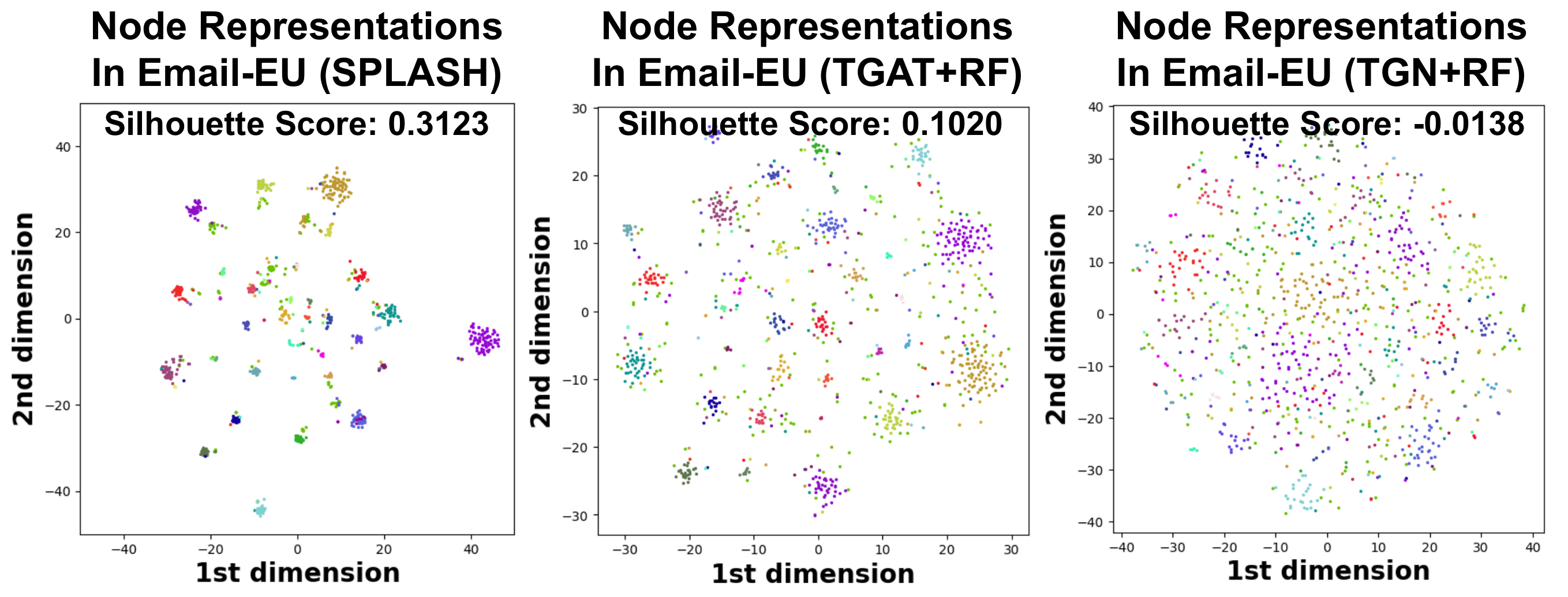} 
    \vspace{-7mm}
    \caption{
    \label{fig:emb_visualzation}
         Node representations produced by \method and two baselines 
         from the Email-EU datasets. 
         The representations are visualized using t-SNE, with colors indicating user properties (classes).
\method produces more distinct and well-separated class representations than the baselines.
    }
\end{figure}

\smallsection{Qualitative Analysis}
\label{qualitative_analysis}
Figure~\ref{fig:case_study} shows the anomaly scores over time predicted by three baselines (DyGFormer+$RF$, FreeDyG+$RF$, TGAT) and \method for a Reddit user (ID: 1292) transitioning from a normal state to an abnormal state.
Note that only \method  accurately detects this transition, appropriately raising the anomaly score in response.

Next, we visualize node representations from the Email-EU dataset obtained by two baselines (TGAT+$RF$, TGN+$RF$) and \method, where each node has a static class as its property. As shown in the t-SNE plots in Figure~\ref{fig:emb_visualzation}, \method produces more cohesive clusters for node representations of the same class and exhibits clearer separation between different classes than the baselines (see also the silhouette scores).




	\section{Conclusion}
\label{sec:conclusion}

In this work, we proposed \method, a simple yet effective method for dynamic node property prediction in CTDG under distribution shifts.
Given a CTDG,
\method enhances its effectiveness by augmenting node features, including those for nodes unseen during training. \method automatically selects the features to augment, ensuring robustness against distribution shifts. Moreover, \method employs \model, a novel lightweight TGNN that uses only MLPs, achieving high accuracy under distribution shifts with efficiency.
Our experiments on three node-property prediction tasks across seven real-world datasets show that, in most cases, \method achieves the highest prediction accuracy among eight TGNNs.



{\small  \smallsection{Acknowledgements}
This work was partly supported by the National Research Foundation of Korea (NRF) grant funded
by the Korea government (MSIT) (No. RS-2024-00406985, 30\%).
This work was partly supported by Institute of Information \& Communications Technology Planning \& Evaluation (IITP) grant funded by the Korea government (MSIT) (No. 2022-0-00157/RS-2022-II220157, Robust, Fair, Extensible Data-Centric Continual Learning, 30\%) 
(No. RS-2024-00438638, EntireDB2AI: Foundations and Software for Comprehensive Deep Representation Learning and Prediction on Entire Relational Databases, 30\%) (No. RS-2019-II190075, Artificial Intelligence Graduate School Program (KAIST), 10\%).}

        \bibliographystyle{IEEEtran}
        
	\bibliography{ref}
        

\end{document}